\documentclass[journal]{IEEEtran}
\ifCLASSINFOpdf
   \usepackage[pdftex]{graphicx}
   \usepackage{amsmath,epsfig,amssymb,subfigure,bm,dsfont}
   \usepackage{multirow}
   \usepackage{epstopdf}
   \usepackage{tabu}
  \usepackage{xcolor}
  \usepackage[ruled,lined,linesnumbered,]{algorithm2e}
  \usepackage{makecell}
\else

\fi

\begin{document}
\title{Generalized Visual Quality Assessment of GAN-Generated Face Images}


\author{Yu Tian, Zhangkai Ni,~\IEEEmembership{Member,~IEEE}, Baoliang Chen, Shiqi Wang,~\IEEEmembership{Senior Member,~IEEE}, 
\\Hanli~Wang,~\IEEEmembership{Senior Member,~IEEE,} 
and Sam Kwong,~\IEEEmembership{Fellow,~IEEE}
\thanks{Yu Tian, Baoliang Chen and Shiqi Wang are with the Department of Computer Science, City University of Hong Kong, Hong Kong 999077 (e-mail:ytian73-c@my.cityu.edu.hk; blchen6-c@my.cityu.edu.hk, shiqwang@cityu.edu.hk).}
\thanks{Zhangkai Ni is with the Department of Computer Science \& Technology, Tongji University, Shanghai 200092, P. R. China (e-mail: eezkni@gmail.com).}
\thanks{Hanli Wang is with the Department of Computer Science \& Technology, Key Laboratory of Embedded System and Service Computing (Ministry of Education), and Shanghai Institute of Intelligent Science and Technology, Tongji University, Shanghai 200092, P. R. China (e-mail: hanliwang@tongji.edu.cn).}
\thanks{Sam Kwong is with the Department of Computer Science, City University of Hong Kong, Hong Kong 999077, and also with the City University of Hong Kong Shenzhen Research Institute, Shenzhen 518057, China (e-mail: cssamk@cityu.edu.hk).}
}

\markboth{
}%
{Shell \MakeLowercase{\textit{et al.}}: Bare Demo of IEEEtran.cls for IEEE Journals}

\maketitle
\begin{abstract}

Recent years have witnessed the dramatically increased interest in face generation with generative adversarial networks (GANs). A number of successful GAN algorithms have been developed to produce vivid face images towards different application scenarios. However, little work has been dedicated to automatic quality assessment of such GAN-generated face images (GFIs), even less have been devoted to generalized and robust quality assessment of GFIs generated with unseen GAN model. Herein, we make the first attempt to study the subjective and objective quality towards generalized quality assessment of GFIs. More specifically, we establish a large-scale database consisting of GFIs from four GAN algorithms, the pseudo labels from image quality assessment (IQA) measures, as well as the human opinion scores via subjective testing. Subsequently, 
we develop a quality assessment model that is able to deliver accurate quality predictions for GFIs from both available and unseen GAN algorithms based on meta-learning. In particular, to learn shared knowledge from GFIs pairs that are born of limited GAN algorithms, we develop the convolutional block attention (CBA) and facial attributes-based analysis (ABA) modules, ensuring that the learned knowledge tends to be consistent with human visual perception. 
Extensive experiments exhibit that the proposed model achieves better performance compared with the state-of-the-art IQA models, and is capable of retaining the effectiveness when evaluating GFIs from the unseen GAN algorithms.
\end{abstract}

\begin{IEEEkeywords} 
Generative adversarial network, domain generalization, meta-learning, face image quality assessment.
\end{IEEEkeywords}

%
\IEEEpeerreviewmaketitle

\section{Introduction}
\label{sec:intro}

\IEEEPARstart{G}{enerative} adversarial network (GAN)~\cite{GoodfellowG14,WGAN217} has achieved remarkable success in various areas such as image generation~\cite{Progan18,StyleGAN19,StyleGAN220}, image restoration~\cite{StarGAN220,LahiriP20}, image quality enhancement~\cite{ni2020towards}, as well as the face images generation~\cite{MaskGAN20,YangA21,InterFaceGAN20}.
In recent years, facial image synthesis and editing have been applied in different areas, such as film, medical aesthetics institutions, and photography technologies. For those specific applications in which the ultimate receiver is human visual system (HVS), face images synthesized by the generative models are expected to be consistent with the perception of HVS. Therefore, the image quality assessment (IQA) models that can predict the perceptual quality of GAN-generated face images (GFIs) are highly desirable. 

\begin{figure}[t]
\centering
\includegraphics[width=1\linewidth]{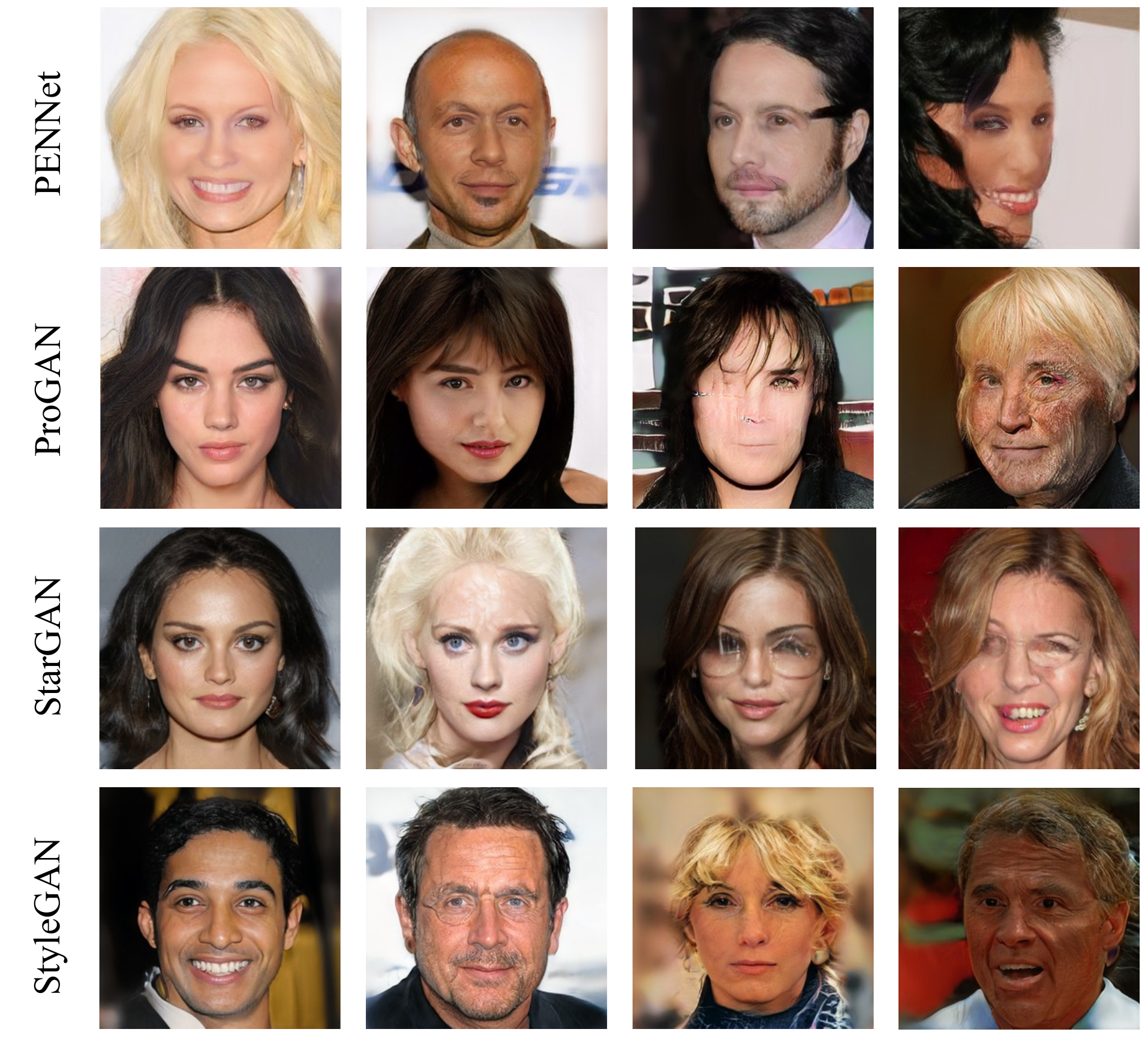}\\
\caption{Illustration of the samples in the proposed GFID.}
\label{fig:sample}
\end{figure}

In the literature, numerous IQA models have been proposed. According to the availability of the reference image, existing IQA models can be divided into three categories: full-reference IQA (FR-IQA) \cite{SSIM04,GMSD14,ding2020image}, reduced-reference IQA (RR-IQA)~\cite{RehmanR12,GolestanehR16}, and no-reference IQA (NR-IQA) \cite{Meta-IQA20,RankIQA17,MaB21,chen2021no}. However, most existing methods are dedicated to natural images with traditional distortion types such as compression, noise and blur. As shown in Fig.~\ref{fig:sample}, GFIs may exhibit substantially different types of distortions, and those distortions on critical face attributes even make GFI look unreal. Therefore, the specific IQA model which predicts the quality of GFIs is in high demand. 


Evaluating the performance of GANs and the quality of GAN-generated images have attracted increasing attention during the past years. The most commonly adopted metrics are Inception score (IS) \cite{IS16} and Frechet Inception distance (FID) \cite{FID17}. IS \cite{IS16} exploits the pre-trained Inception~\cite{Inception16} classifier to extract the class probability of each generated image. Then the Kullback–Leibler (KL) divergence between the probability of this image and the marginal distribution is regarded as the performance of the GAN model. FID \cite{FID17} compares activations between real data and generated data in an intermediate pooling layer of Inception~\cite{Inception16}. It is worth mentioning that IS and FID can only measure the overall performance of GAN model instead of each individual image. Furthermore, Gu \textit{et al.} \cite{Gu20} proposed three GAN-generated images quality assessment (GIQA) methods from different perspectives, and the GIQA model with the best performance is based on Gaussian mixture model (GMM), termed as GMM-GIQA. Unfortunately, GMM-GIQA mainly focuses on the probability of the generated data in the distribution of real data (i.e., training data) while ignoring the perceptual characteristics relevant to HVS. 

\begin{figure}[t]
\centering
\includegraphics[width=1\linewidth]{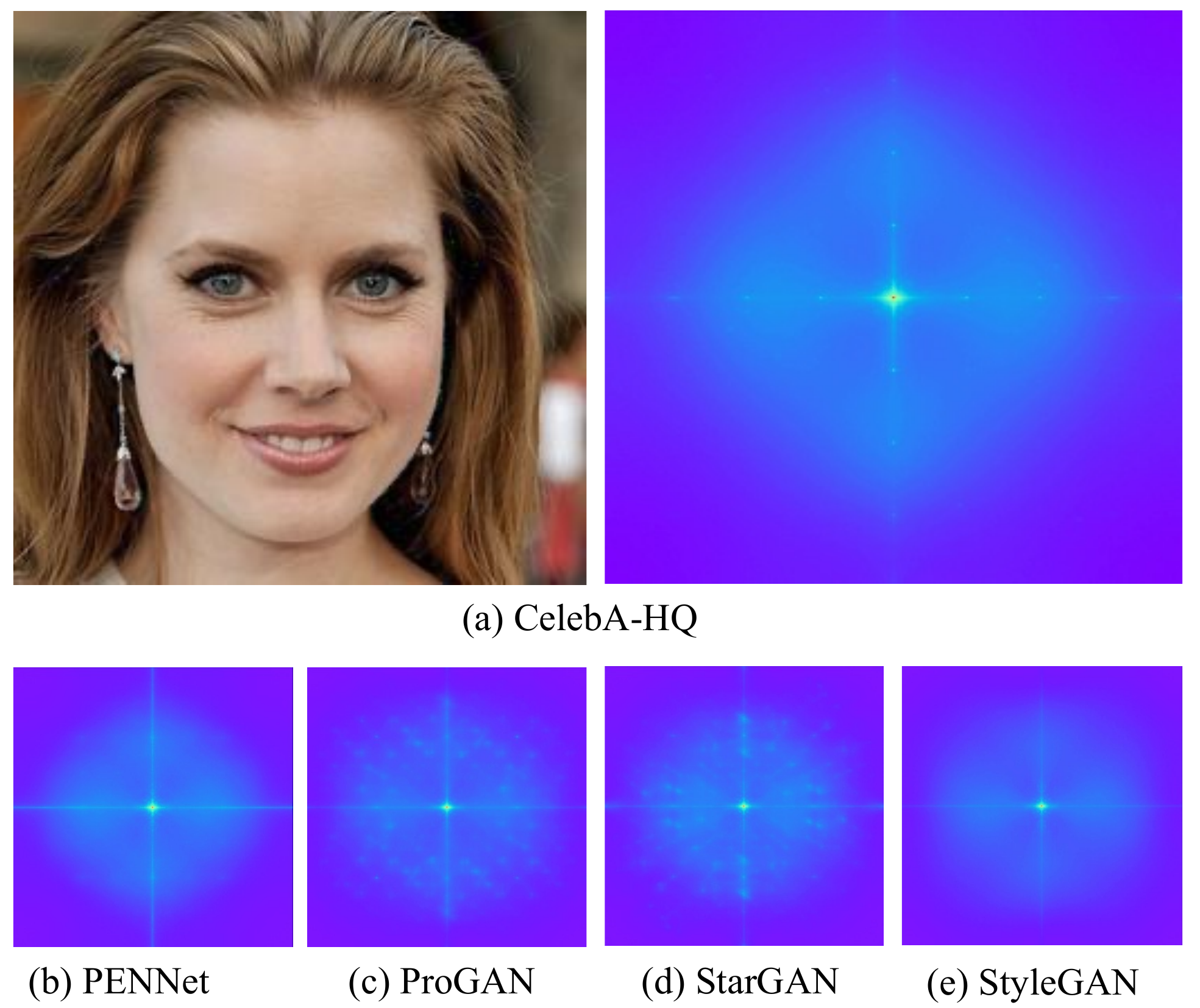}\\
\caption{The average spectra for the real face images in CelebA-HQ database~\cite{Progan18} and  face images generated by PENNet \cite{PENNet19}, ProGAN \cite{Progan18}, StarGAN \cite{StarGAN220} and StyleGAN \cite{StyleGAN19}, respectively. Each average spectra is generated by averaging the frequency spectra of over 5,000 face images.}
\label{fig:frequency}
\end{figure}

In this paper, we make one of the first attempts to conduct subjective and objective studies on the quality assessment of GFIs. Regarding the dataset, we collect a large number of GFIs from four different GAN algorithms to establish metric-labeled training data and human-labeled testing data.  
Eventually, our database contains 200,000 training image pairs and 2,000 testing images. Regarding the objective model, given the fact that there exists large
domain gaps among GFIs from different GAN algorithms, which can be observed in the form of the average spectra as shown in Fig.~\ref{fig:frequency}, we aim to develop an objective model that is robust to different GAN algorithms even in the unseen domain.  
To this end, we develop the GFIs quality assessment measure by associating the meta-learning optimization strategy with quality-relevant features, which enhances the generalization capability and improves the prediction performance. 
In summary, the main contributions of this paper are summarized as follows,
\begin{itemize}
  \item 
     We establish the large-scale GFIs database (GFID) for GFIs quality assessment, which contains the training image pairs associated with relative quality information and the testing images with human-annotated mean opinion scores (MOSs).
\item
  We develop a new GFIs quality assessment measure that is robust to different GAN algorithms with high generalization capability. 
  More specifically, we devise the convolutional block attention (CBA) and facial attributes-based analysis (ABA) modules to extract the quality-relevant features over GFIs of each source GAN, and adopt the meta-learning strategy to mine the shared features representation across GFIs from different GAN algorithms.
   \item
    We perform extensive experiments, which provide useful evidence that  our model is more consistent with the perception of the HVS when assessing GFIs quality compared to state-of-the-art IQA methods. In particular, our model is able to achieve good generalization performance on unknown GANs.
\end{itemize}

\section{Related works}
\subsection{Image Quality Assessment}
IQA is a long-standing research topic regarding how to provide an objective measurement of the image quality that is consistent with the HVS. Most FR-IQA models have achieved promising performance by directly comparing the distorted image and its corresponding pristine version, such as the structural similarity (SSIM) index~\cite{SSIM04}, the visual saliency-induced (VSI) index \cite{VSI14}, the visual information fidelity (VIF) metric~\cite{VIF06}. NR-IQA models, however, have attracted increasing attentions in recent years due to the limited available pristine images.
Early NR-IQA models are developed based on knowledge-driven strategy \cite{SheikhNo05,FerzliA09}, aiming at designing a quality-aware feature descriptor. Subsequently, a regression model, e.g., support vector regression (SVR), maps the feature representation into a quality score. Based on the assumption that the natural scene statistics (NSS) characterized from natural images can govern the quality of natural images and the NSS will be destroyed in the presence of distortion, Moorthy \textit{et al.} \cite{MoorthyB11} predicted the image quality by characterizing the NSS of distorted images in the wavelet domain. Mittal \textit{et al.} \cite{BRISQUE12} exploited the normalized luminance coefficients to map image features in the spatial domain into the quality score. Instead of only measuring the NSS regularities, Mittal \textit{et al.} \cite{NIQE13} further proposed a naturalness image quality evaluator (NIQE) based on quality-aware statistical features.  

Recently, deep-learning-based NR-IQA methods have been proposed due to the powerful feature representation capability. Ma \textit{et al.} \cite{MEON18} proposed an end-to-end quality assessment model based on multi-task learning, which consists of a quality assessment sub-network and a distortion identification sub-network. Zhang \textit{et al.} \cite{DBCNN20} proposed a deep bilinear model for synthetically and authentically distorted images. The quality score can be obtained by bilinearly pooling two feature representations extracted from two CNN models. Su \textit{et al.} \cite{HyperIQA20} combined quality prediction with content understanding and proposed a self-adaptive IQA model. To improve the generalization of NR-IQA models, Zhu \textit{et al.} \cite{Meta-IQA20} used the deep meta-learning approach to develop a novel NR-IQA model called Meta-IQA. 
However, the limitation for deep-learning-based NR-IQA methods is that the lack of human-labeled training data makes the model training difficult. To address this problem, Liu \textit{et al.} \cite{RankIQA17} ranked images according to distortion levels of distorted images and then used the ranked images to train the siamese network. Finally, they fine-tuned the model to transform the task from learning-to-rank to the quality score regression. Ma \textit{et al.} \cite{dipIQ17} applied RankNet \cite{RankNet05} to learn the quality relationship between the quality-discriminable image pairs. Chen \textit{et al.} \cite{chen2021no} explored the unified distribution regularization on the feature space and developed an unsupervised domain adaptation based quality assessment model for screen content images.

\subsection{Domain Generalization} 
Domain generalization (DG) aims to learn a model that can generalize to unseen data distributions by training on one or more data distributions. Previous studies on DG are mainly based on four different strategies. The first is to use the discrepancy measurement, e.g., maximum mean discrepancy (MMD) \cite{LiD18,YangM13}, to learn domain-invariant features. Such domain-invariant features can work well in both source and unseen target domains. The second resorts to the feature disentanglement, which is based on the assumption that each domain can be presented as the combination of the domain-specific component and the domain-agnostic component. Thus, the model learned from the domain-agnostic component \cite{KhoslaU12,LiD17} can achieve better generalization performance. The third is based on data augmentation, which synthesizes additional training data to enhance the robustness of the models to target domains \cite{ShankarG18,ZhouD21}. The last is to find minima during training, called optimization strategy. The most popular optimization strategy is meta-learning \cite{LiL18,SohM20,Meta-IQA20}, which improves a conventional learning algorithm to improve the robustness of the models on unseen domains. The fundamental idea behind meta-learning is that simulating the real train/test data shift during training improves the generalization ability. In this paper, our approach applies meta-learning to build domain-shift batches at each episode to optimize it for learning more robust face representations.
\section{The Proposed Database}
The proposed GFID database is established for the purpose of developing a reliable quality assessment model. As such, the database includes a training dataset and a testing dataset.
\begin{table}
\renewcommand{\arraystretch}{1.6}
\tabcolsep0.3cm
\centering
\caption{The details of the GFID testing dataset.}
    \begin{tabular}{c|c|c|c}
    \hline
    \hline
    Model & Year  &  \# of Images & Task \\
    \hline
    StarGAN~\cite{StarGAN220} & 2018  & 469   & Image-to-image translation \\
    \hline
    ProGAN~\cite{Progan18} & 2018  & 515   & Image generation \\
    \hline
    PENNet~\cite{PENNet19} & 2019  & 527   & Image inpainting \\
    \hline
    StyleGAN~\cite{StyleGAN19} & 2019  & 489   & Image generation \\
    \hline
    \hline
    \end{tabular}%
  \label{tab:gan}%
\end{table}%
\subsection{Training Dataset}
\label{sec:TrainingData}
We first obtain GFIs from different GAN algorithms. To cover various types of distortions in different application scenarios, we select four GAN models, including the pyramid-context encoder network (PENNet)~\cite{PENNet19} for image inpainting/restoration, the GAN using a progressive training methodology (ProGAN) \cite{Progan18}, the style-based GAN (StyleGAN) \cite{StyleGAN19} for unconditional image generation, and the StarGAN \cite{StarGAN220} for multi-domain image-to-image translation. Each GAN model generates 15,000 face images forming four subdatasets. It is worth mentioning that we use the pre-trained versions of the above-mentioned GAN models which are trained on the large-scale CelebFaces attributes dataset (i.e., CelebA-HQ~\cite{Progan18}) including over 30,000 high-quality celebrity images. Furthermore, to acquire the quality levels of the GFIs, we adopt the pseudo-labels based approach, as human annotation of such a large quantity of images is tedious and expensive. Considering that information content weighted SSIM (IW-SSIM) \cite{IWSSIM11} has been widely used to evaluate the quality of GFIs in image inpainting and restoration tasks, and it can reflect image degradations at low-level feature by comparing the restored image and the corresponding pristine version, we select the IW-SSIM values as the pseudo-MOS values of GFIs generated by PENNet. For GFIs from ProGAN, StyleGAN, and StarGAN, the corresponding pristine version of each GFI is unavailable. Herein, the GMM-GIQA model \cite{Gu20} is applied to produce the pseudo-MOS values. More specifically, we first use the Inception-v3 network \cite{Inception16} to extract the feature representations of real face images in CelebA-HQ as suggested in~\cite{Gu20}. Subsequently, we apply the method provided by the authors~\cite{Gu20} to build the GMM model to capture the data distribution from the feature representations. Finally, the probabilities of GFIs computed from the GMM can be used as the pseudo-MOS values of GFIs generated by ProGAN, StyleGAN, and StarGAN.

Inspired by the promising transferability of pair-wise relationship in IQA, we establish a pair-wise training dataset with the relative quality ranking automatically generated by pseudo-MOS values of two images within each image pair. To further mitigate the over-fitting problem caused by distinguishing image pairs with ambiguous quality, we collect quality-discriminable image pairs from each subdataset. More specifically, we first sort all GFIs within each subdataset according to their corresponding pseudo-MOS values from high to low. Then, we divide the sorted GFIs into three quality levels, of which each quality level contains 5,000 GFIs. Subsequently, we sample the top 50 and the worst 50 GFIs from the first and the third quality levels, respectively. Within the 100 images, each image is paired with 500 GFIs randomly sampled from the second quality level. Finally, our training dataset contains 200,000 image pairs with relative quality rankings in total, of which each subdataset consists of 50,000 quality-discriminable image pairs.
\subsection{Testing Dataset}
Testing data contains a total of 2,000 GFIs. More specifically, we initially collect around 3,000 face images generated by PENNet \cite{PENNet19}, ProGAN \cite{Progan18}, StyleGAN \cite{StyleGAN19}, and StarGAN \cite{StarGAN220}. Subsequently, we apply the semantic segmentation network~\cite{BiSeNet18} to analyze the generated images and refine the selection according to the richness of semantic constituents. This step ensures the diversity of generated image content and quality. Finally, 2,000 GFIs are selected, and we resize them to a fixed resolution of 256$\times$256. The details of the testing data can be found in Table \ref{tab:gan} and the sampled images are shown in Fig.~\ref{fig:sample}. The subjective experiment is conducted to collect human opinion scores of the 2,000 testing GFIs. Such scores are treated as the ground-truth labels to examine whether IQA models are well correlated with human perception.
\subsubsection{Subjective Test}
\begin{figure}[t]
\centering
\includegraphics[width=0.9\linewidth]{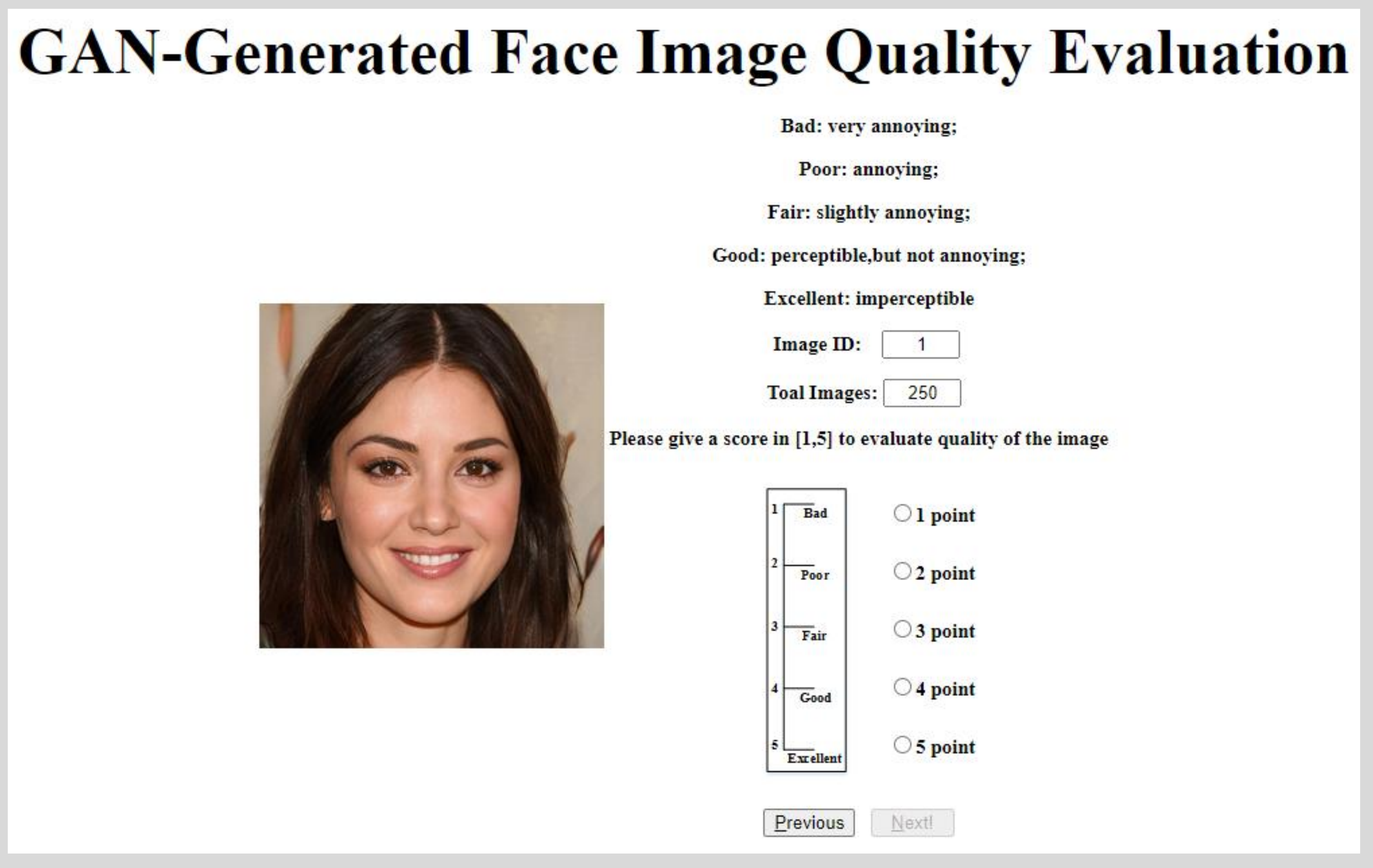}\\
\caption{A screenshot of the user interface in the subjective test.}
\label{fig:testsystem}
\end{figure}

In order to reduce the effect of the viewer fatigue, we randomly and non-overlappingly divide the entire 2,000 GFIs into eight sessions. As specified in ITU-R BT.500-13~\cite{ITU}, each session is evaluated by at least 15 subjects. Before starting the testing session, we provide a sample set of GFIs to ensure each subject has fully understood how to rate the quality scores of GFIs. For the large-scale image database, based on the recommendations of ITU~\cite{ITU}, the single stimulus approach with 5-category discrete scales is used for the subjective experiment.
From the highest perceptual quality to the lowest perceptual quality, the impairment scales are classified as ``Excellent", ``Good", ``Fair", ``Poor", ``Bad". During the subjective experiment, subjects are asked to choose the subjective value in the range between 1 to 5 according to the impairment scale of the image. The higher value indicates the better perceptual quality. The graphical user interface 
is shown in Fig.~\ref{fig:testsystem} and we conduct the experiment in a well-controlled laboratory environment.

\subsubsection{Subjective Score Processing}
After obtaining the subjective scores directly rated by subjects, we further process the subjective results to generate the MOS of each GFI. As suggested in~\cite{ITU, ni2017esim}, we denote $S_{ijk}$ as the opinion score of $i$-th image evaluated by subject \emph{j} in \emph{k}-th testing session, where $i={1,\dots,20},j={1,\dots,250}$ and $k={1,\dots,8}$. We convert the $S_{ijk}$ to \emph{Z}-score per session as follows:
\begin{equation}
\begin{split}
\mu_{jk}&=\frac{1}{N_{k}} \sum_{i=1}^{N_{k}},\\
\sigma_{jk}&=\sqrt{\frac{1}{N_{k}-1}\sum_{i=1}^{N_{k}}(S_{ijk}-\mu _{jk}) ^{2}} ,\\
Z_{ijk} &= \frac{S_{ijk}-\mu_{jk}}{\sigma _{jk}},
\end{split}
\end{equation}
where $N_{k}$ is the number of subjects in the test session $k$.

Assuming that Z-scores assigned by a subject follow a standard Gaussian, 99\% values will fall in the range between -3 and +3. Therefore, Z-scores can be re-scaled to the range of [0,100] by the following linear mapping,
\begin{equation}
  \tilde{Z}_{ijk} = \frac{100(Z_{ijk} +3)}{6}.  
\end{equation}

Finally, the MOS of the GFI $j$ in $k$-th testing session, which is denoted as $MOS_{jk}$, can be obtained as:
\begin{equation}
    MOS_{jk}=\frac{1}{M_{k}} \sum_{i=1}^{M_{k}} \tilde{Z}_{ijk}, 
\end{equation}
where $M_{k}$ is the number of remaining subjects in the session $k$ after the subject rejection.

\begin{figure*}[t]
\centering
\includegraphics[width=1\linewidth]{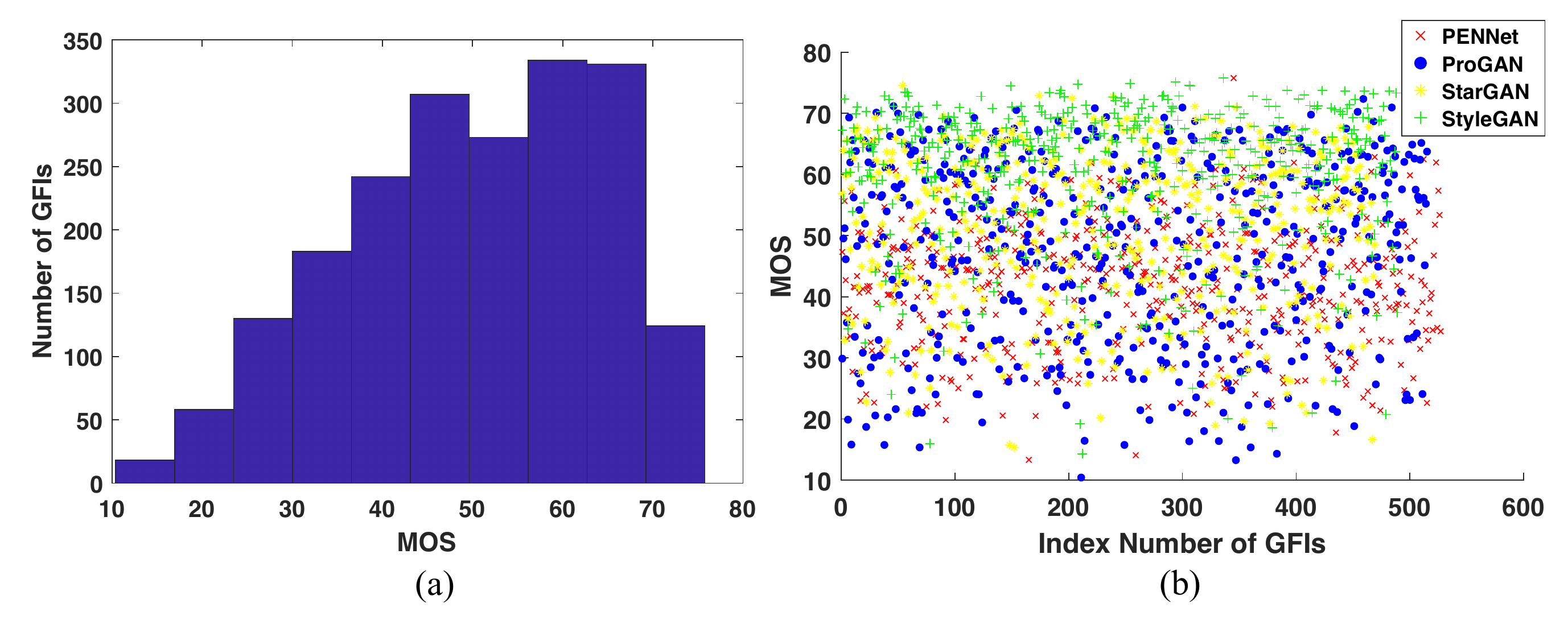}\\
\caption{The subjective scores (i.e., MOS) distribution in the testing dataset of our proposed GFID. (a) Histogram of MOS; (b) Scatter plot for different GAN algorithms.}
\label{fig:testscore}
\end{figure*}

To analyze the reliability of the proposed testing dataset, we visualize the MOS distribution of all tested GFIs via the histogram and the scatter plots. As shown in 
Fig.~\ref{fig:testscore}, one can easily observe that the MOS values of all GFIs are distributed between 10 and 80, which implies the perceptual quality of GFIs in our dataset has spanned a wide range of visual quality from severely annoying to imperceptible with a good separation. 

\begin{figure*}[t]
\centering
\includegraphics[width=1\linewidth]{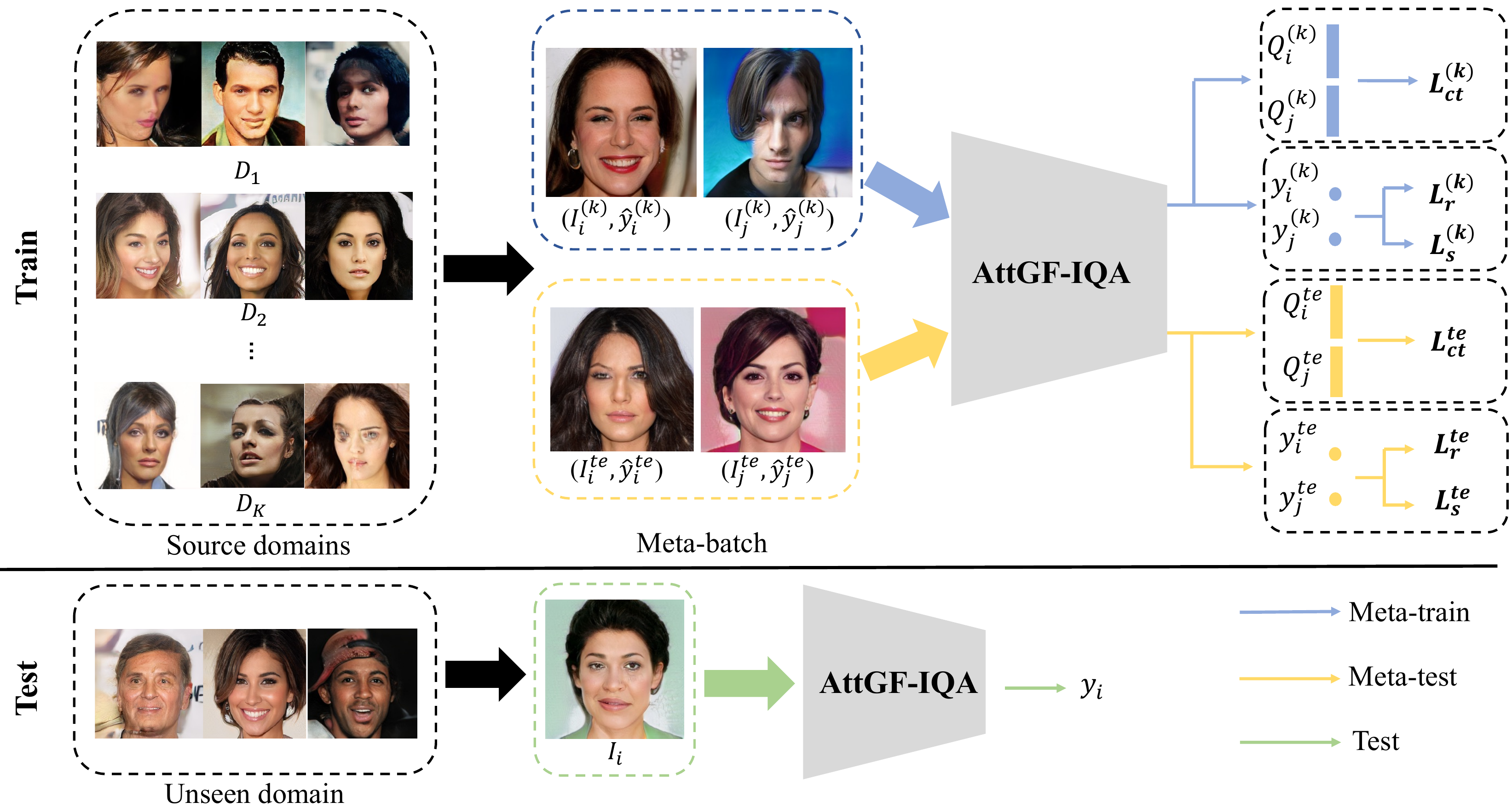}\\
\caption{The overview of the meta-learning strategy. The purpose is to improve the generalization ability of the proposed model. The ``Train" part is the training process of our model. We split source domains into a meta-batch to simulate the real train/test data shift at every training iteration. The meta-batch includes meta-train and meta-test sub-batches, and all sub-batches work together for the optimization updates of our model. Finally, the trained model is tested on the unseen domain.}
\label{fig:meta}
\end{figure*}
\section{Proposed Method}

\subsection{Overview}
Let $\mathcal{I}$ denote an input space consisting of GFIs and $\mathcal{\hat{Y}}$ is the output space representing the pseudo-MOS values of samples in $\mathcal{I}$. Given $K$ training (source) domains generated by $K$ types of GAN methods, $\mathcal{D}=\left\{\mathcal{D}_{k} \mid k=1, \cdots, K\right\}$, where $\mathcal{D}_{k}=\left\{\left(I_{i}^{(k)}, \hat{y}_{i}^{(k)}\right),\left(I_{j}^{(k)}, \hat{y}_{j}^{(k)}\right)\right\}_{i\ne j}$ denotes the $k$-th training domain containing GFIs pairs and the pseudo-MOS values, we can infer the relative quality rankings by comparing the pseudo-MOS values of two images within image pairs. More specifically, we set the relative quality ranking to 1 if the pseudo-MOS value $\hat{y}_{i}^{(k)}$ of the image $I_{i}^{(k)}$ is higher than the pseudo-MOS value $\hat{y}_{j}^{(k)}$ of the image $I_{j}^{(k)}$ and 0 otherwise.
Our goal is to learn a generalizable predictive function $h$ via the pairwise learning-to-rank approach within $K$ training domains to achieve a minimum prediction error on an unseen testing domain $\mathcal{D}_{\text {unseen }}$:
\begin{equation}
\min _{h} \mathbb{E}_{(I,MOS) \in \mathcal{D}_{\text {unseen}}}[\ell(h(I), MOS)],
\end{equation}
where $MOS$ is the human-annotated score of the image $I$, $\mathbb{E}$ is the expectation, and $\ell(\cdot, \cdot)$ is the loss function. In particular, the  $\mathcal{D}_{\text {unseen }}$ is generated by the GAN models which is exactly unseen during the training phase, thus the joint distributions between the training and testing sets are dramatically different.
The prediction model is required to perform well on source domains as well as the unseen domains. Herein, we aim to measure the generalization capability of our IQA models with the blind assumption on the specific generative model. Such scenario plays a more practical role in real applications.   

\begin{figure*}[t]
\centering
\includegraphics[width=1\linewidth]{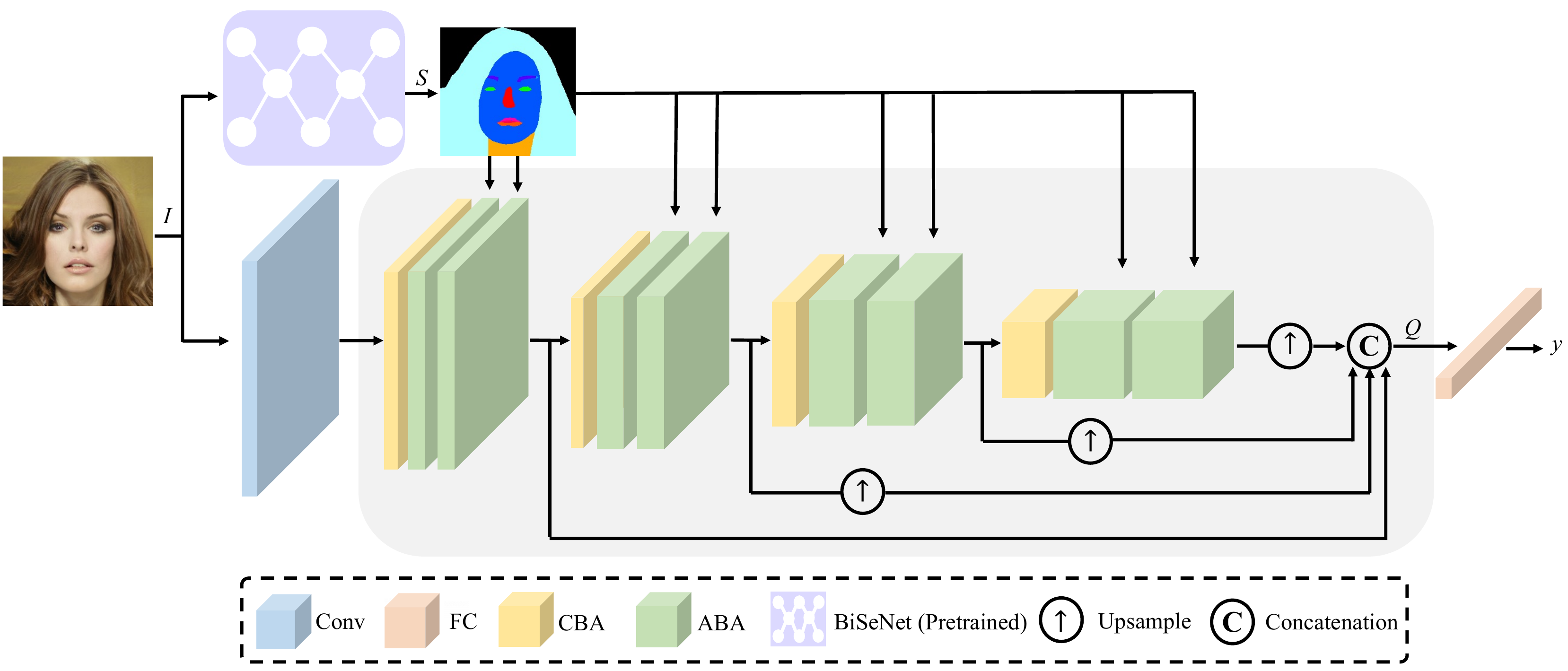}\\
\caption{The architecture of the proposed AttGF-IQA, including the CBA module and the ABA module.}
\label{fig:framework}
\end{figure*}

To this end, we first propose the CBA and ABA modules to capture the 
intrinsic and perceptual aware features delicately. In particular, the CBA module aims to mine the visual attention region of the face image and the ABA module attempts to extract perceptual feature representations from different facial attributes in visual attention regions. Subsequently, the meta-learning approach is adopt to learn the prior knowledge of the HVS perception shared by different domains.
 The overview of the meta-learning strategy and the architecture of the proposed model are shown in Fig. \ref{fig:meta} and Fig.~\ref{fig:framework}, respectively, and we will elaborate the design details of each module in the following subsections.

\begin{figure}[t]
\centering
\includegraphics[width=1\linewidth]{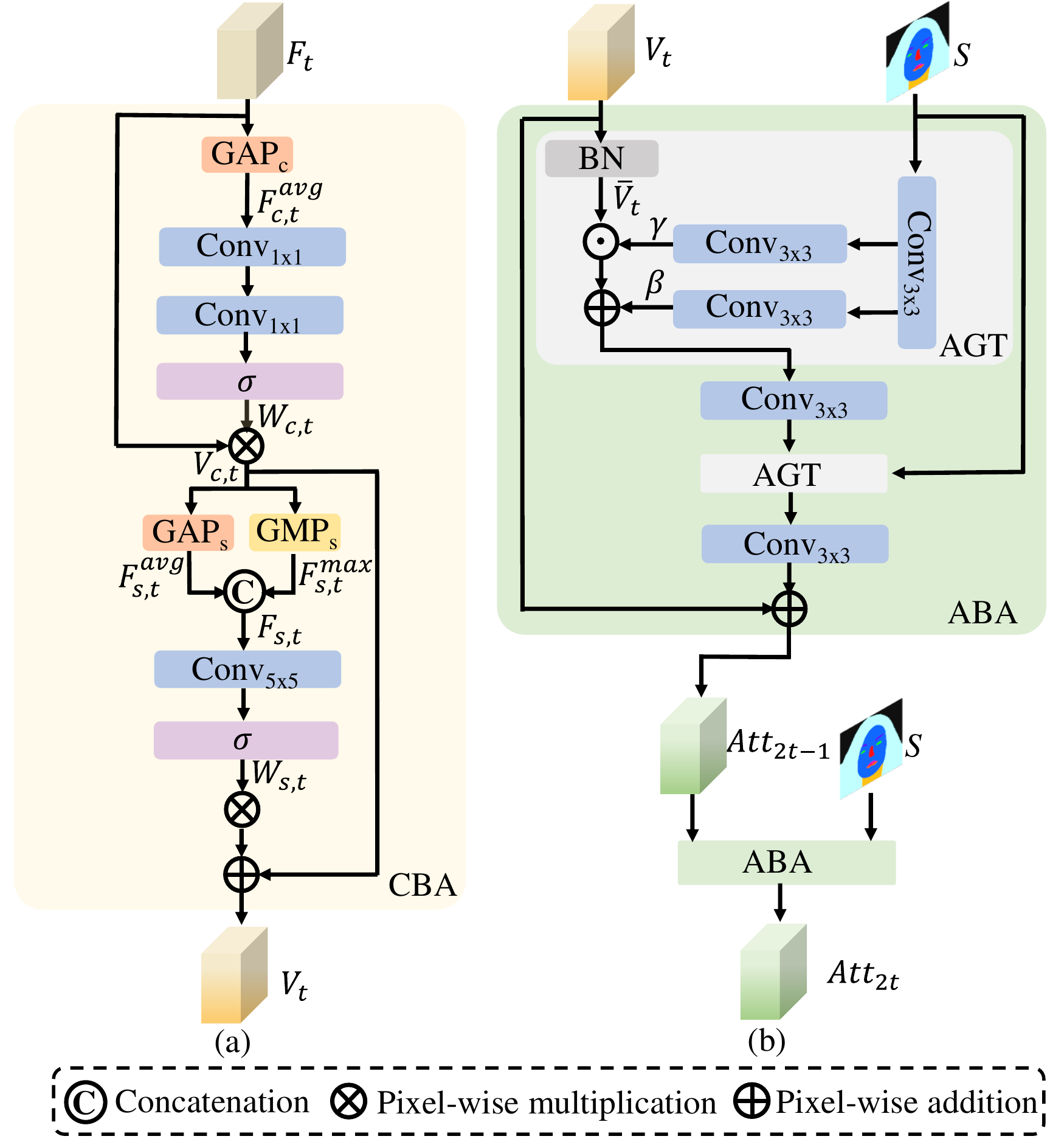}\\
\caption{The detailed structure of (a) CBA and (b) ABA.}
\label{fig:module}
\end{figure}
\subsection{Convolutional Block Attention Module}

Inspired by human visual attention mechanism, the attention module is designed for high task-relevant feature learning \cite{CBAM18,U-GAT-IT20}.
Considering that channel and spatial attentions have different meanings for the whole image, i.e., ``what" is important and ``where" is informative, respectively, 
we use the CBA module to capture the salient feature by aggregating the channel and spatial attention features. The structure of CBA is shown in Fig.~\ref{fig:module}(a). To be specific, the input of the $t$-th CBA module in Fig. \ref{fig:framework} is denoted as $F_{t}\in \mathbb{R}^{C\times H\times W}$. We first take global average pooling operation (denoted as $\text{GAP}_{c}$) to aggregate spatial features of each channel to obtain the global feature descriptor (denoted as $F_{c,t}^{avg}\in \mathbb{R}^{C\times 1\times 1}$). Then we feed it into two $1\times 1$ convolutions with stride 1 (denoted as $\text{Conv}_{1\times 1}$) and apply a sigmoid activation layer (denoted as $\sigma$) to produce the channel weighting map $W_{c,t}$,
\begin{equation}
\begin{split}
    W_{c,t}&=\sigma(\text{Conv}_{1\times 1}(\text{Conv}_{1\times 1}(\text{GAP}_{c}(F_{t}))))\\
    &=\sigma(\text{Conv}_{1\times 1}(\text{Conv}_{1\times 1}(F_{c,t}^{avg}))).
\end{split}
\end{equation}
 It is worth mentioning that in order to reduce the model complexity, the dimension of the feature $F_{c,t}^{avg}\in \mathbb{R}^{C\times 1\times 1}$ should be reduced as $\mathbb{R}^{C/r\times 1\times 1}$ using $1$-th $\text{Conv}_{1\times 1}$, where $r$ is the reduction radio. After obtaining the channel weighting map, we multiply it with the input features $F_{t}$ to generate the channel attention feature $V_{c,t}$,
\begin{equation}
    V_{c,t}= W_{c,t} \otimes F_{t},
\end{equation}
where the symbol ``$\otimes$" denotes elemental-wise multiplication.

Regarding the spatial attention, the max pooling and average pooling layers (denoted as $\text{GMP}_{s}$ and $\text{GAP}_{s}$) are further employed along the channel axis of $V_{c,t}$, respectively. Subsequently, we
concatenate the outputs 
to generate the spatial feature descriptor denoted as $F_{s,t}\in \mathbb{R}^{2\times H\times W}$. Finally, the spatial attention map $W_{s,t}$ can be acquired with a convolutional layer (denoted as $\text{Conv}_{5\times 5}$) and a sigmoid activation layer followed, which is given by,
\begin{equation}
\begin{split}
    W_{s,t}&=\sigma (\text{Conv}_{5\times 5}([\text{GMP}_{s}(V_{c,t}),\text{GAP}_{s}(V_{c,t})]))\\
    &=\sigma (\text{Conv}_{5\times 5}(\left[ F_{s,t}^{max}, F_{s,t}^{avg}\right ] ))\\
    &=\sigma (\text{Conv}_{5\times 5}(F_{s,t})),
\end{split}
\end{equation}
where $[\cdot,\cdot ]$ represents the channel-wise concatenation, $F_{s,t}^{max}$ and $F_{s,t}^{avg}$ are the outputs of $\text{GMP}_{s}$ and $\text{GAP}_{s}$. 

The final visual attention features $V_{t}\in \mathbb{R}^{C\times H\times W}$ is given by,
\begin{equation}
    V_{t} = W_{s,t} \otimes V_{c,t} + V_{c,t}.
\end{equation}
As shown in Fig.~\ref{fig:framework},
the CBA module is incorporated before every two ABA modules, thus the salient information can be exploited at different scales.

\subsection{Attributes-Based Analysis Module}

A face image can be described by several meaningful attributes, such as hair, eye, nose, mouth, skin, which play important roles for the quality evaluation of GFIs \cite{KumarD11}. Along this vein, we transform the visual attention features extracted by CBA into attributes-based features and predict the quality score by aggregating different attributes.
The structure of our ABA module is shown in Fig.~\ref{fig:module}(b). More specifically, the ABA module requires two inputs, including the output of the previous module and a face segmentation map. For example, the inputs of the first ABA module after $t$-th CBA are the visual attention feature ${V}_{t}$ and a face segmentation map $S$ generated by a pre-trained face parsing network BiSeNet~\cite{BiSeNet18}. Herein, to incorporate the semantic information into the attention features, the conditional normalization is adopted, grounded on the philosophy that different attributes should own different feature statistics \cite{InterFaceGAN20}. In particular, we first use batch-normalization layer (denoted as BN) to normalize the attention feature ${V}_{t}$ as follows, 
\begin{equation}
    \mu_c = \frac{1}{NHW}\sum_{nyx}V_{n,c,y,x},
\end{equation}
\begin{equation}
    \sigma_c = \sqrt{\frac{1}{NHW}\left (   \sum_{nyx}V_{n,c,y,x}^{2}-\mu_{c}^{2} \right )} ,
\end{equation}
where $H$, $W$ and $C$ are the height, width, and the number of channels in $V$. $N$ is the number of training samples in a batch. The normalized ${V}_{t}$ is denoted as $\bar{V}_{t}$. Following the $\bar{V}_{t}$, inspired by SPADE~\cite{SPADE19}, we propose an attributes-guided transform (AGT) sub-module to explore the guidance information of the segmentation map and generate affine transformation parameters $(\gamma, \beta)$ for $\bar{V}_{t}$.
As shown in Fig.~\ref{fig:module}(b), our AGT consists of two convolutional layers with their kernels set as $3\times 3$ for the $\gamma, \beta$ extraction. 
Subsequently, we compute the attributes-guided activation feature $T$ by denormalizing the feature $\bar{V}$ according to the affine transformation parameters $(\gamma, \beta)$,
\begin{equation}
    T = \gamma \otimes \bar{V} + \beta,
\end{equation}
where the dimensions of $\gamma$ and $\beta$ are the same as $\bar{V}$. Then, we use $3 \times 3$ convolution after every AGT sub-module to obtain the attribute-based features. For the given input ${V}_{t}$, the output of the ABA module is denoted as $Att_{2t-1}$. We concatenate attribute-based features at different layers of the network to generate the multi-scale attribute-based feature $Q$. Finally, the predicted quality score $y$ is defined as:
\begin{equation}
\begin{split}
    y &= \text{FC}(Q)\\    
    & = \text{FC}(\left[ Att_{2},Att_{4},Att_{6},Att_{8}\right]),
\end{split}
\end{equation}
where $Att_{i}$ presents the output of $i$-th attribute-guided transform in Fig.~\ref{fig:framework}. $\text{FC}$ is the fully connected layer.

\subsection{Meta-Learning Strategy}
As discussed above, the main goal of our IQA model is to achieve a high genralization capability even on the face images generated by unseen GAN models. To this end, the shareable and transferable features should be extracted from the source domains, aiming to mitigate the domain gaps caused by different GAN models. Following this vein, the meta-learning strategy is utilized in our method.
To begin with, we first
split a meta-batch from source domains at each iteration and a meta-batch contains several sub-batches. Every sub-batch is further sampled as a meta-test dataset (denoted as $D_{te}$) and the rest as meta-train datasets (denoted as $D_{tr}$). The training process of the proposed model on a sub-batch is shown in the top of the Fig.~\ref{fig:meta} that comprises two stages: (1) meta-train on $D_{tr}$; (2) meta-test on $D_{te}$. During the meta-train phase, we denote the image pair as ($I_{i}^{(k)},\hat{y}_{i}^{(k)}$) and ($I_{j}^{(k)},\hat{y}_{j}^{(k)}$), where $I_{i}^{(k)}$ and $I_{j}^{(k)}$ are the $i$-th and $j$-th images from $k$-th $D_{tr}$, respectively. $\hat{y}_{i}^{(k)}$ and $\hat{y}_{j}^{(k)}$ are the corresponding pseudo-MOS values of $I_{i}^{(k)}$ and $I_{j}^{(k)}$, respectively. 
The quality scores of $I_{i}^{(k)}$ and $I_{j}^{(k)}$ predicted by our proposed model are $y_{i}^{(k)}$ and $y_{j}^{(k)}$. Then, we can compute the rank loss $L_{r}^{(k)}$ according to the predicted probability of $I_{i}^{(k)}$ is better than $I_{j}^{(k)}$ \cite{focal17}. The predicted probability $p_{ij}^{(k)}$ is formulated as:
\begin{equation}
\begin{split}
        p_{ij}^{(k)}&=\left\{\begin{matrix}
  &\frac{\text{exp}(y_{i}^{(k)}-y_{j}^{(k)})}{1+\text{exp}(y_{i}^{(k)}-y_{j}^{(k)})} ,&\text{if}~\hat{y}_{i}^{(k)}\ge \hat{y}_{j}^{(k)},\\
  &1-\frac{\text{exp}(y_{i}^{(k)}-y_{j}^{(k)})}{1+\text{exp}(y_{i}^{(k)}-y_{j}^{(k)})}, &\text{otherwise}.
\end{matrix}\right.
        \end{split}
\end{equation}

The rank loss $L_{r}^{(k)}$ aims to minimize the difference between the predicted probability and the ground-truth ranking order.
\begin{equation}
    L_{r}^{(k)}=\frac{1}{N}\sum_{i,j}^{N}-(1-p_{ij}^{(k)})^{\gamma }\text{log}(p_{ij}^{(k)}),
\end{equation}
where $\gamma$ is the focusing parameter for adjusting the importance of easy samples and hard samples. $N$ is the number of training image pairs in $k$-th $D_{tr}$.

\begin{algorithm}
\SetKwInOut{Input}{input}
\SetKwInOut{Init}{Init}

\Input{Source domains (GANs) $D=\left\{ D_{1},D_{2},\dots ,D_{K} \right\}$.}
\Init{The AttGF-IQA model $f_{\theta}$ parametrized by $\theta$, learning rate $\beta$.}
\BlankLine

\For{iteration = 1,2, $\dots$}
{
    \tcp{For a meta-batch}
    { Initialize the gradient $g_{\theta}$ as 0;}\\
     \For{each $D_{te}$ in D}
     {
     \tcp{For a sub-batch}
     {Sample remaining $K-1$ domains (GAN models) as $D_{tr}$;}\\
     {
     \textbf{Meta-train:}\\
     Compute the meta-train loss on the parameter $\theta$\\
     by:\\
     $L_{tr} = \sum_{k=1}^{K-1}(\lambda_{0}L_{r}^{(k)} +\lambda_{1}L_{ct}^{(k)}+\lambda_{2}L_{s}^{(k)})$ \\
     Update the parameters $\theta$ by:\\
     {
     $\theta^{'} = Adam(L_{tr},\theta)$ on $D_{tr}$}\\
     }
     \textbf{Meta-test:}\\
     Compute the meta-test loss on the updated\\
     parameter $\theta^{'}$ by:\\
     $L_{te} = \lambda_{3}L_{r}^{te} +\lambda_{4}L_{ct}^{te}+\lambda_{2}L_{s}^{te}$\\
     Aggregate gradient:\\
     $g_{\theta} \gets g_{\theta} +\bigtriangledown _{\theta }L_{tr} +\bigtriangledown _{\theta }L_{te}$\\
     }
    update $\theta \gets \theta - \beta\frac{1}{N}g_{\theta} $
}
\caption{Meta-learning optimization}
\label{alg:aa}
\end{algorithm}

Although the meta learning strategy adopted here can align the distributions among different domains, the discrimination capability of the learned model is still not ensured \cite{zhu2020deep}. To address this issue, the center loss \cite{center16, chen2021no} is further introduced, aiming to align the conditional distributions of different domains. In particular, we learn two renewable prototypes (centers) as the conditions of the ranking features, then the intra-class distances of the ranking features are minimized.
More specifically, the ranking feature $R_{ij}^{(k)}$ is the difference between the features $Q_{i}^{(k)}$ and $Q_{j}^{(k)}$ before the last fully connected layer: 
\begin{equation}
    R_{ij}^{(k)} = Q_{i}^{(k)}-Q_{j}^{(k)}.
\end{equation}

Thus, the center loss is computed as
\begin{equation}
\label{eq:metatrainct}
\begin{split}
    L_{ct}^{(k)}=&\frac{1}{N}\sum_{i,j}^{N}  ( \delta (\hat{y}_{i}^{(k)}< \hat{y}_{j}^{(k)} ) {\left \| R_{ij}^{(k)}- c_{0}\right \| }_{2}^{2} \\
    +&\delta (\hat{y}_{i}^{(k)} \ge \hat{y}_{j}^{(k)} ) {\left \| R_{ij}^{(k)}- c_{1}\right \| }_{2}^{2}),
\end{split}
\end{equation}
where $\delta(condition)=1$ if the condition is satisfied, and $\delta(condition)=0$ otherwise. $c_{0}$ and $c_{1}$ are the learned class centers of the ranking features $R_{ij}^{(k)}$. $N$ is the number of training image pairs.

For the images collected from the face image inpainting model, their corresponding pseudo-MOS values are computed by the FR-IQA objective model IW-SSIM. Those objective scores reflect the similarity between real and generated images in terms of low-level features (e.g., image structure and texture). Thus, we add the score regression loss when training data contains the inpainted face images. Supposing GFIs pairs of $\alpha$-th source domain is collected from face image inpainting model, the score regression loss is defined as:
\begin{equation}
 L_{s}^{(k)} = \begin{cases}
   \frac{1}{M}\sum_{m}^{M}(\hat{y}_{m}^{(\alpha)}-{y}_{m}^{(\alpha)}),&\text{ if } k=\alpha,   \\
   0,&\text{otherwise},
\end{cases}
\end{equation}
where $M$ is the number of training images in the $\alpha$-th source domain.

The loss function of the meta-train can be summarized as follows,
\begin{equation}
\label{eq:metatrainloss}
    L_{tr} = \sum_{k=1}^{K-1}(\lambda_{0}L_{r}^{(k)} +\lambda_{1}L_{ct}^{(k)}+\lambda_{2}L_{s}^{(k)}),
\end{equation}
where $K-1$ is the number of meta-train databases during the meta-train procedure. $\lambda_{0}$, $\lambda_{1}$ and $\lambda_{2}$ are the weighting factors of $L_{r}^{(k)}$, $L_{ct}^{(k)}$ and $L_{s}^{(k)}$, respectively. 

After meta-training, the model with the knowledge learned from meta-train datasets $D_{tr}$ needs to remain effective when testing on the meta-test dataset $D_{te}$. 
Analogously, we also compute the meta-test loss trained on the meta-test dataset $D_{te}$. As such,
\begin{equation}
\label{eq:metatestloss}
    L_{te} = \lambda_{3}L_{r}^{te} +\lambda_{4}L_{ct}^{te}+\lambda_{2}L_{s}^{te},
\end{equation}
where $\lambda_{3}$ and $\lambda_{4}$ is the weighting factors of $L_{r}^{te}$ and $L_{ct}^{te}$, respectively. 

We define the AttGF-IQA model represented by a parametrized function $f_{\theta}$ with parameters $\theta$. 
Following the similar strategies as suggested in \cite{Meta-IQA20}, we update the parameters $\theta$ via Stochastic Gradient Descent (SGD). The whole meta-leaning procedure is summarized in Algorithm \ref{alg:aa}.

\begin{table*}[htbp]
\renewcommand{\arraystretch}{1.4}
\tabcolsep0.2cm
  \centering
  \caption{Generalization performance comparisons of the proposed AttGF-IQA and state-of-the-art NR-IQA models on the unseen domain in the GFID testing dataset, where the best, the second-best and the third-best results are boldfaced in red, blue, and black, respectively.}
    \begin{tabular}{c|c|ccccccccc|c}
    \hline
    \hline
    Unseen & Measures & NIQE  & FRIQUEE & MEON  & NIMA  & Meta-IQA & dipIQ & UNIQUE&RankIQA & GMM-GIQA & AttGF-IQA \\
     Domain & ~ & \cite{NIQE13}  & \cite{FRIQUEE17} & \cite{MEON18}  & \cite{NIMA18} & \cite{Meta-IQA20} & \cite{dipIQ17} & \cite{UNIQUE21} &~\cite{RankIQA17}& ~\cite{Gu20} & (Ours) \\ 
    \hline
    \multicolumn{1}{c|}{\multirow{2}[1]{*}{ StyleGAN}} & SRCC  & 0.0919 & 0.1315 & -0.0514 & 0.0396 & 0.0955 & 0.2284 & 0.0463&\textbf{0.4658} & \textcolor{blue}{\textbf{0.5101}} &  \textcolor{red}{\textbf{0.7326}}\\
          & PLCC  & 0.0396 & 0.1257 & -0.0053 & 0.0120 & 0.1762 & 0.2301 & 0.0694&\textcolor{blue}{\textbf{0.6044}} &\textbf{0.5743} & \textcolor{red}{\textbf{0.7538}} \\
    \hline
    \multicolumn{1}{c|}{\multirow{2}[1]{*}{ StarGAN}} & SRCC  & -0.0483 & 0.2531 & 0.0168 & 0.2711 & 0.2700 & 0.0676 & 0.1055 & \textbf{0.3275}& \textcolor{blue}{\textbf{0.3676}} & \textcolor{red}{\textbf{0.6459}} \\
          & PLCC  & -0.0640 & 0.2655 & -0.0215 & 0.2794 & 0.2814 & 0.0599 & 0.1017&\textbf{0.3214} & \textcolor{blue}{\textbf{0.3537}} & \textcolor{red}{\textbf{0.6495}} \\
    \hline
    \multicolumn{1}{c|}{\multirow{2}[1]{*}{ProGAN}} & SRCC  & 0.0092 & 0.1832 & 0.0834 & 0.0567 & 0.1434 & 0.1640 & 0.0286 &\textbf{0.4215}& \textcolor{blue}{\textbf{0.4726}} & \textcolor{red}{\textbf{0.7728}} \\
          & PLCC  & -0.0064 & 0.1654 & 0.1215 & 0.0395 & 0.1583 & 0.2162 & 0.0453&\textbf{0.4595} & \textcolor{blue}{\textbf{0.4914}} & \textcolor{red}{\textbf{0.7791}} \\
    \hline 
    \multicolumn{1}{c|}{\multirow{2}[1]{*}{PENNet}} & SRCC  & -0.1160 & 0.3073 & 0.1338 & 0.1757 & 0.2661 & -0.0633 & 0.3182 &\textbf{0.3436}& \textcolor{blue}{\textbf{0.3809}} & \textcolor{red}{\textbf{0.6952}} \\
          & PLCC  & -0.1561 & 0.3283 & 0.1610 & 0.1850 & 0.2323 & -0.0479 & 0.3289 &\textbf{0.3806}& \textcolor{blue}{\textbf{0.3906}} & \textcolor{red}{\textbf{0.6718}} \\
    \hline
    \hline
    \end{tabular}%
  \label{tab:generalization}%
\end{table*}%

\begin{table}
\renewcommand{\arraystretch}{1.6}
\tabcolsep0.4cm
  \centering
  \caption{Performance comparisons of the proposed model and the state-of-the-art FR IQA models on GFIs generated by PENNet in the GFID testing dataset. The proposed model is trained with GFIs pairs of ProGAN, StyleGAN, and StarGAN in the GFID training dataset.}
    \begin{tabular}{l|rr}
    \hline
    \hline
         Methods & \multicolumn{1}{l}{SRCC} & \multicolumn{1}{l}{PLCC} \\
    \hline
    PSNR & 0.4891&0.4776	\\
    SSIM~\cite{SSIM04}  & 0.3218&	0.3358 \\
    IW-SSIM~\cite{IWSSIM11}  & 0.5781&0.5629	 \\
    VSI~\cite{VSI14}  & 0.4493&0.3084	 \\
    VIF~\cite{VIF06}  &0.4046 &	0.4036 \\
    AttGF-IQA & \textbf{0.6952}    & \textbf{0.6718} \\
    \hline
    \hline
    \end{tabular}%
  \label{tab:friqa}%
\end{table}%

\begin{figure*}
\label{fig:scatter}
\centering
\begin{minipage}[b]{0.47\linewidth}
  \centering
  \centerline{\includegraphics[width=1\linewidth]{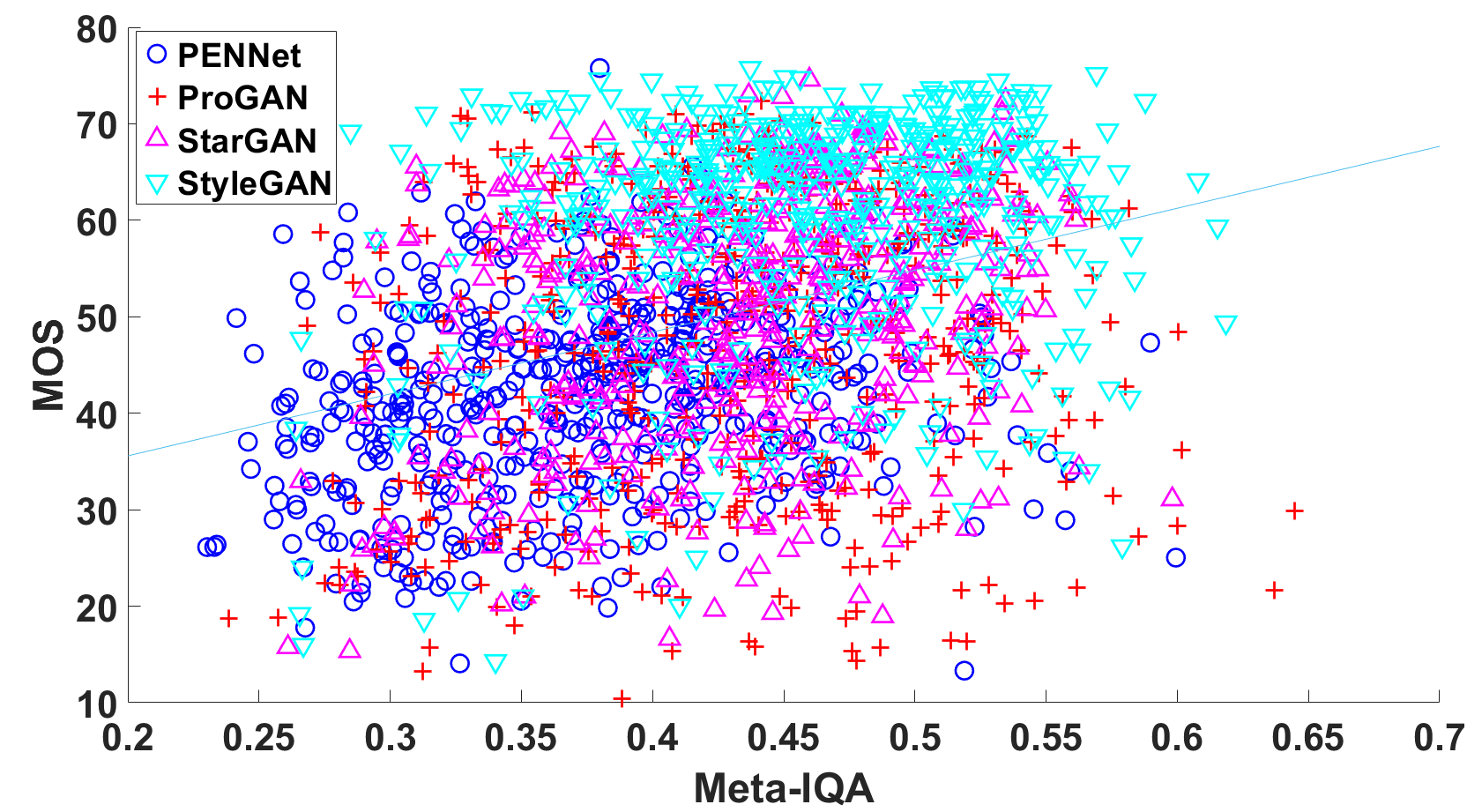}}
  \centerline{(a)} 
\end{minipage}
\begin{minipage}[b]{0.47\linewidth}
  \centering
  \centerline{\includegraphics[width=1\linewidth]{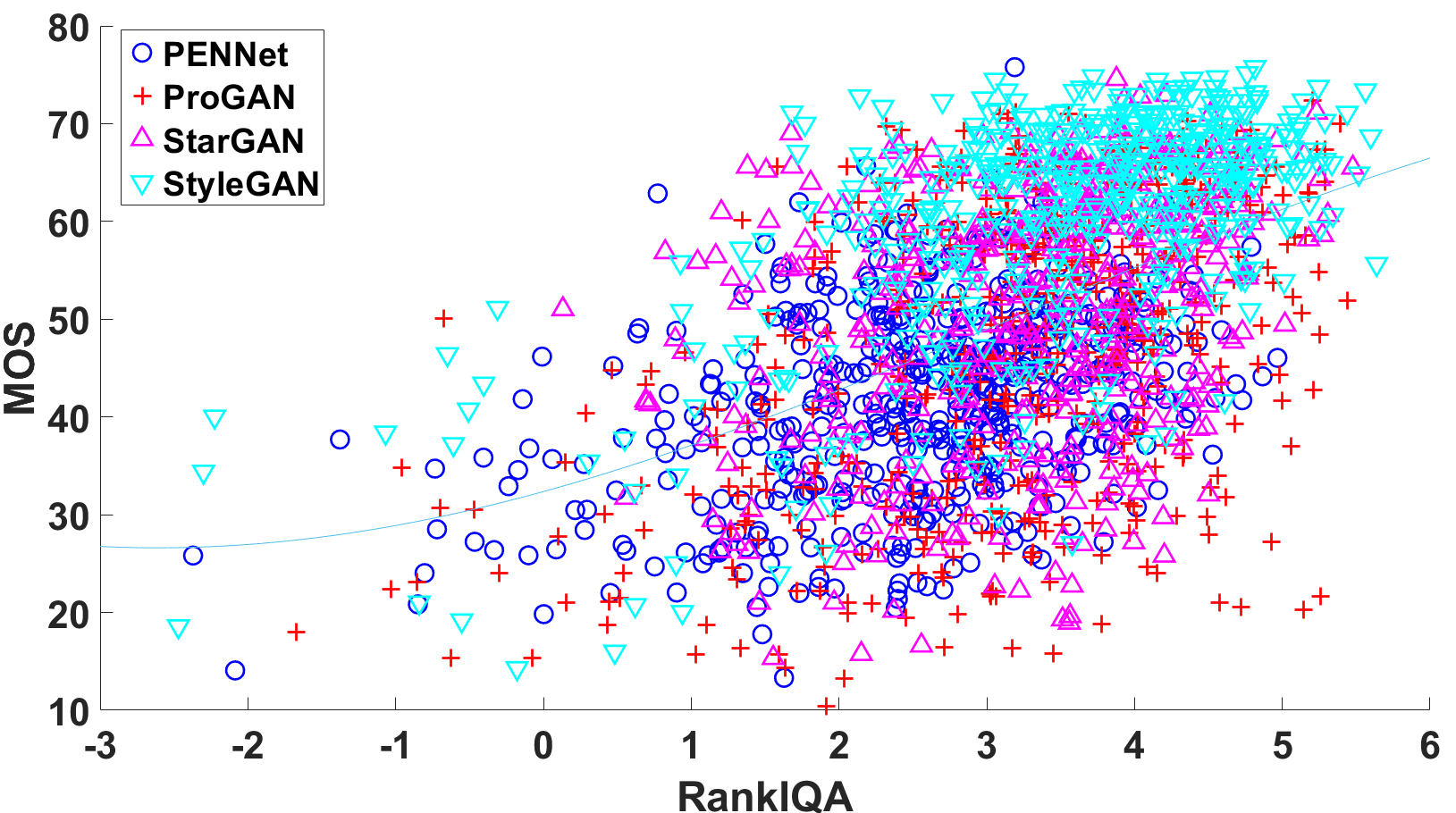}}
  \centerline{(b)} 
\end{minipage}
\begin{minipage}[b]{0.46\linewidth}
  \centering
  \centerline{\includegraphics[width=1\linewidth]{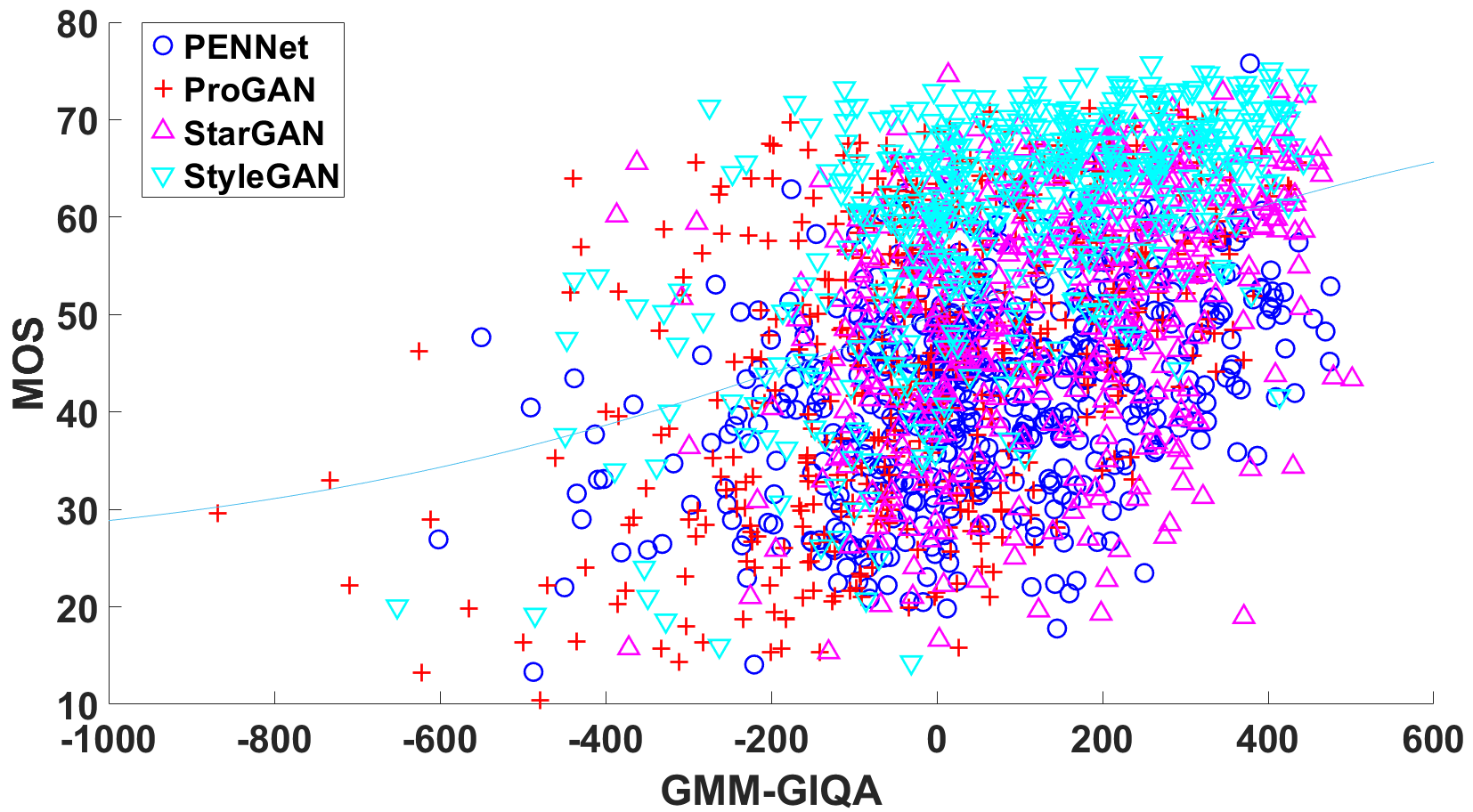}}
  \centerline{(c) }
\end{minipage}
\begin{minipage}[b]{0.46\linewidth}
  \centering
  \centerline{\includegraphics[width=1\linewidth]{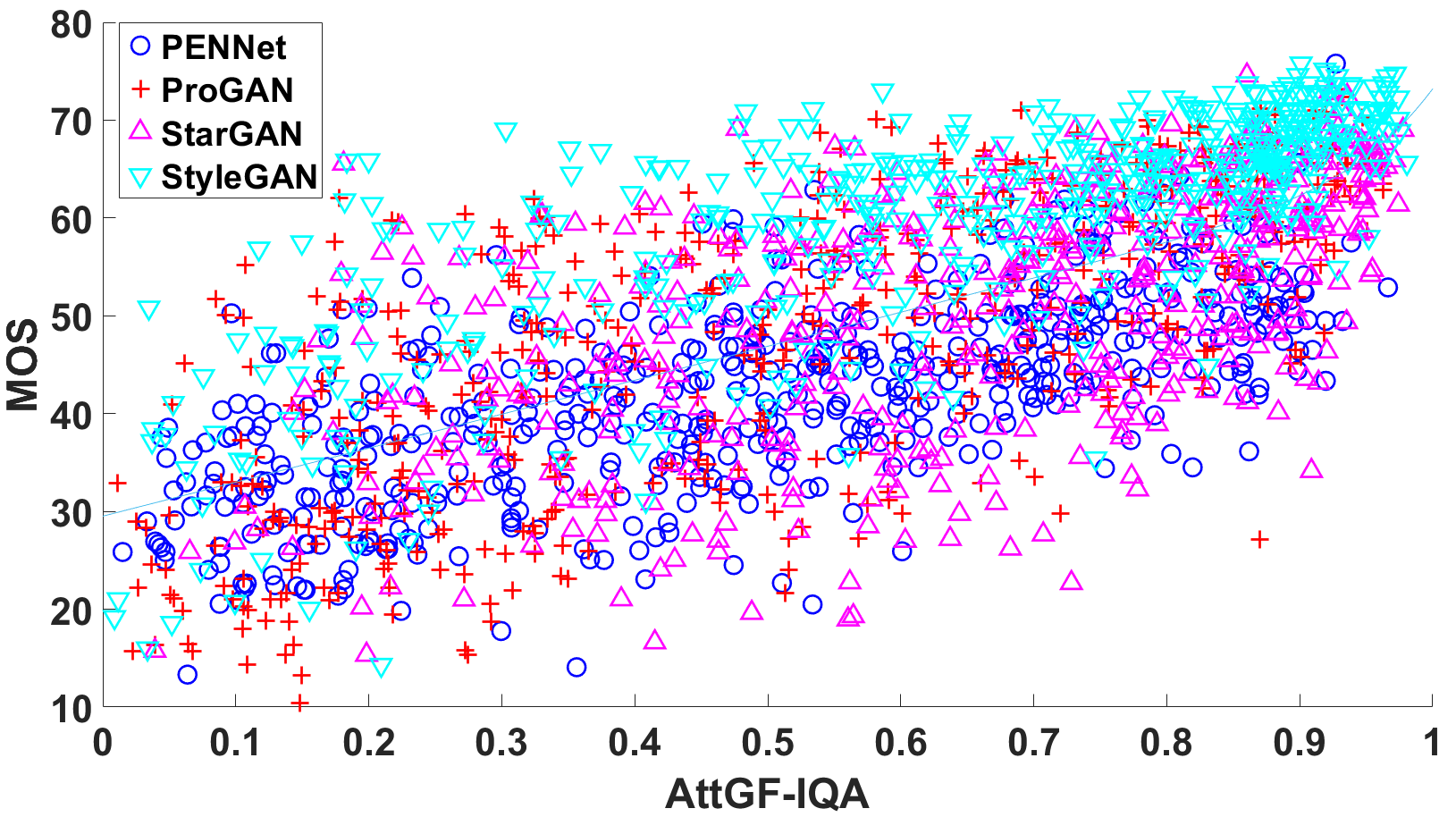}}
  \centerline{(d) }
\end{minipage}
\caption{Scatter plots of the subjective scores (i.e., MOS values) versus the objective scores computed by NR-IQA models on the GFID testing dataset. (a) Meta-IQA; (b) RankIQA; (c) GMM-GIQA; (d) the proposed AttGF-IQA trained on GFIs pairs of StarGAN, StyleGAN, and PENNet in the GFID training dataset.
}
\label{fig:scatter}
\end{figure*}
\begin{table}
\renewcommand{\arraystretch}{1.4}
\tabcolsep0.2cm
  \centering
  \caption{SRCC and PLCC results of the proposed model when training on source domains in the GFID training dataset and testing on both source and unseen domains in the GFID testing dataset.}
    \begin{tabular}{l|llll}
    \hline
    \hline
    Unseen&\multicolumn{4}{l}{~~~~~~~~~~~~~~~~~~~~~~~~~~SRCC}\\
        \cline{2-5}
         Domain & StyleGAN & StarGAN & ProGAN& PENNet \\
    \hline
       StyleGAN &   0.7326    &  0.5967        &0.7200    &0.6856  \\
    StarGAN &   0.7045    & 0.6459      &   0.7576    &  0.7428\\
    ProGAN &    0.7015   & 0.6409      &   0.7728    & 0.7733 \\
          PENNet &   0.6858    &   0.6096    &    0.7520   & 0.6952 \\
    \hline
    \hline
    Unseen&\multicolumn{4}{l}{~~~~~~~~~~~~~~~~~~~~~~~~~~PLCC}\\
        \cline{2-5}
                 Domain & StyleGAN & StarGAN & ProGAN& PENNet \\
    \hline

          StyleGAN &  0.7538     & 0.5789     &  0.7194       & 0.6748\\
    StarGAN &  0.7815     &  0.6184     &    0.7791   &  0.7310\\
     ProGAN &  0.7717     &  0.6599     &  0.7638     &0.7704 \\
           PENNet &  0.7359     &  0.5718     &  0.7408     & 0.6718 \\
    \hline
    \hline
    \end{tabular}%
  \label{tab:individualtest}%
\end{table}%

\section{Experiments}
\subsection{Implementation Details}

The proposed GFID database consists of training GFIs pairs and testing GFIs which are collected from four different GAN algorithms, and we treat GFIs generated by each GAN model as an independent domain. Therefore, there are four domains in our proposed GFID database. To simulate the cross-domain scenario, three domains are selected as the source domains for training, and the remaining one domain is used for testing which is strictly unseen in the training phase.


We implement our model by PyTorch~\cite{pytorch}. The inputs of our network are resized to $256\times256\times3$, and the batch size is set as 10. We apply the Adam~\cite{Adam} optimizer with a learning rate of $1e-5$ and a weight decay of $5e-5$. The learning rate is reduced by 5 after every 10 epochs and the maximum epoch is 100. The weighting parameters $\lambda_{0}$, $\lambda_{1}$, $\lambda_{2}$, $\lambda_{3}$, and $\lambda_{4}$ in Eqn. (\ref{eq:metatrainloss}) and Eqn. (\ref{eq:metatestloss}) are set as 10, 0.01, 1, 10, and 0.01, respectively.
\subsection{Evaluation Criteria}
To compare the performance of our proposed model with other IQA models, we employ two common criteria, i.e., Spearman Rank Correlation Coefficient (SRCC) and Pearson Linear Correlation Coefficient (PLCC). The higher values of SRCC and PLCC, the better prediction performance the model can achieve. As suggested in the \cite{VQEG, ni2018gabor}, before computing those correlation coefficients, we adopt a nonlinear logistic mapping function to map the dynamic range of the predicted scores from various IQA metrics onto a common scale. The mapped score $s_{i}$ of $i$-th face image can be computed as 
\begin{equation}
\label{eq:nonlinear}
    s_{i}=\beta_{1}\left ( \frac{1}{2}-\frac{1}{e^{\beta_{2}(y_{i}-\beta_{3})}}  \right ) +\beta_{4}y_{i}+\beta_{5},
\end{equation}
where $y_{i}$ is the predicted scores of $i$-th face image from an IQA metric. $\beta_{1}$, $\beta_{2}$, $\beta_{3}$, $\beta_{4}$ and $\beta_{5}$ are to be determined by minimizing the sum of squared errors between the mapped score $s_{i}$ and the subjective score.

\subsection{Performance Comparisons}
To explore the generalization of the proposed model on the unseen domain, we select several NR-IQA metrics for performance comparison, including GMM-GIQA~\cite{Gu20}, NIQE \cite{NIQE13}, FRIQUEE~\cite{FRIQUEE17}, MEON~\cite{MEON18}, NIMA~\cite{NIMA18}, Meta-IQA~\cite{Meta-IQA20}, dipIQ~\cite{dipIQ17}, UNIQUE~\cite{UNIQUE21}, where GMM-GIQA is an IQA model specifically for GIQA tasks. It is worth mentioning that the pre-trained versions of the competing models provided by the corresponding authors are used for performance comparison except for GMM-GIQA. For GMM-GIQA, the GMM used in all testing experiments is the same as the GMM built for the generation of pseudo-MOS in our proposed training dataset. 

Table~\ref{tab:generalization} shows the performance of the proposed model and different NR-IQA models when testing on the unseen domain in the GFID testing dataset. 
It is worth mentioning that the proposed AttGF-IQA is trained on three source domains only in the GFID training dataset.
 The experimental results show that the proposed AttGF-IQA trained on available source domains can perform well on the unseen domain, and its prediction performance outperforms all competing models. The main reason is that most models are designed for evaluating the quality of natural images whose distortions differ widely from GAN-generated distortions. In particular, GAN-generated distortions are related to the network architectures of generative models, which are diverse and difficult to predict. Besides, the features extractor of GMM-GIQA is the pre-trained Inception-v3 network \cite{Inception16} designed for image classification tasks, ignoring the influence of human visual perception. To further demonstrate the effectiveness of our proposed model specifically for evaluating restored images, we also compare the proposed method with several state-of-the-art FR-IQA models, including the peak signal-to-noise ratio (PSNR), SSIM \cite{SSIM04}, IW-SSIM \cite{IWSSIM11}, VSI \cite{VSI14}, VIF \cite{VIF06}. Table \ref{tab:friqa} documents the performance results of all models evaluated on GFIs generated by PENNet in the GFID testing dataset, where AttGF-IQA is the proposed model trained on GFIs pairs of StyleGAN, StarGAN, and ProGAN in the GFID training dataset. From the results, the proposed AttGF-IQA still acquire the best performance without the reference information of pristine images. Besides, from the results of Table \ref{tab:generalization} and Table \ref{tab:friqa}, we notice that although the GFID training dataset uses the quality scores predicted by GMM-IQA and IWSSIM as the pseudo-MOS, the proposed AttGF-IQA can achieve better performance than GMM-IQA and IWSSIM on the GFID testing dataset. This is because the proposed AttGF-IQA learns from quality-discriminable image pairs with relative quality rankings such that it can avoid the overfitting problem caused by overconfidence in the pseudo-MOS of each GFI.

 Table~\ref{tab:individualtest} shows the results when evaluating the models on both seen and unseen domains. In particular, each time we select on GAN algorithm as the unseen domain, and train the proposed model on the corresponding three source domains in the GFID training dataset. Given the trained model, we evaluate the performance on each domain of the testing data. From the experimental results, we can find that those models are still able to deliver promising performance for both seen and unseen domains when using the testing data with ground-truth MOS for evaluation. However, we can also observe relatively low SRCC and PLCC values on the quality assessment of GFIs generated by StarGAN. The reason is that most distortions in GFIs generated by StarGAN appear in facial contour and global spatial information, and the proposed framework fails to extract global image descriptor and reflect the correlation between facial attributes in the GFI. In Fig.~\ref{fig:scatter}, we further show the scatter plots of the MOS values against the objective scores as predicted by the RankIQA, GMM-GIQA, and AttGF-IQA, where the AttGF-IQA is trained on GFIs pairs of PENNet, StyleGAN, and StarGAN in the GFID training dataset, and the objective scores are the predicted results of GFIs of all GAN models in the GFID testing dataset. As shown in Fig.~\ref{fig:scatter}, compared with Meta-IQA, RankIQA, and GMM-GIQA, the proposed AttGF-IQA can better align with the MOS values. It implies the proposed AttGF-IQA is more consistent with visual quality of GFIs.

\begin{table}
\renewcommand{\arraystretch}{1.4}
\tabcolsep0.2cm
  \centering
  \caption{Comparison of SRCC and PLCC of the proposed framework using different optimization strategies on unseen domains in the GFID testing dataset.}
    \begin{tabular}{l|rrrr}
    \hline
    \hline
    \multirow{3}*{Methods}&\multicolumn{4}{l}{~~~~~~~~~~~~~~~~~~~~~~~SRCC}\\
    \cline{2-5}
    &\multicolumn{4}{l}{~~~~~~~~~~~~~~~~~~~Unseen Domain}\\
    \cline{2-5}
    & \multicolumn{1}{l}{StyleGAN} & \multicolumn{1}{l}{StarGAN} & \multicolumn{1}{l}{ProGAN} & \multicolumn{1}{l}{PENNet} \\
    \hline
    s-StyleGAN&--~~~~~&0.4512&0.6271& 0.4817\\
    s-StarGAN&0.4900&--~~~~~&0.5726&0.4231\\
    s-ProGAN&0.6222&0.5215&--~~~~~&0.6582\\
    s-PENNet&0.6309&0.4912&0.7052&--~~~~~\\
    w/o meta-learning &   0.7026    &    0.5826   &   0.7304    &0.6392  \\
    AttGF-IQA &   \textbf{0.7326}    &   \textbf{0.6459}    &   \textbf{0.7728}    &  \textbf{0.6952}\\
    \hline
    \hline
    \multirow{3}*{Methods}&\multicolumn{4}{l}{~~~~~~~~~~~~~~~~~~~~~~~PLCC}\\
    \cline{2-5}
    &\multicolumn{4}{l}{~~~~~~~~~~~~~~~~~~~Unseen Domain}\\
    \cline{2-5}
    & \multicolumn{1}{l}{StyleGAN} & \multicolumn{1}{l}{StarGAN} & \multicolumn{1}{l}{ProGAN} & \multicolumn{1}{l}{PENNet} \\
    \hline
    s-StyleGAN&--~~~~~&0.4585&0.6328& 0.4723\\
    s-StarGAN&0.5457&--~~~~~&0.5938&0.4232\\
    s-ProGAN&0.6929&0.5296&--~~~~~&0.6477\\
    s-PENNet&0.7022&0.5381&0.7244&--~~~~~\\
    w/o meta-learning &  0.7450     &   0.5921    &   0.7353    & 0.6330 \\
    AttGF-IQA &    \textbf{0.7538}   &  \textbf{0.6495}     &    \textbf{0.7791}   & \textbf{0.6718} \\
    \hline
    \hline
    \end{tabular}%
  \label{tab:ab_meta}%
\end{table}%
\subsection{Ablation Studies}
In this subsection, we conduct a series of ablation experiments to confirm the contribution of meta-learning strategy and different components of the proposed model on assessing GFIs quality. 

\subsubsection{Contribution of Meta-Learning Strategy}
To investigate the contribution of meta-learning on the performance of our proposed model, we perform the ablation study based on different optimization strategies. Table~\ref{tab:ab_meta} lists SRCC and PLCC results of the model evaluated on the unseen domain in the GFID testing dataset when using different optimization strategies. In this table, \textbf{s-StyleGAN}, \textbf{s-StarGAN}, \textbf{s-ProGAN}, and \textbf{s-PENNet} represent models trained with GFIs pairs of StyleGAN, StarGAN, ProGAN, and PENNet in the GFID training dataset, respectively. From the results, one can easily observe that the proposed models trained on a single source domain have performance degradation at different degrees. This phenomenon reveals that there exists a domain gap between GFIs generated by different generative models. Moreover, the three source domains are merged as the training data, and we evaluate the model performance without meta-learning strategy. In particular, \textbf{w/o meta-learning} is the model trained with GFIs pairs of three corresponding source domains in the GFID training dataset and directly optimized by an Adam optimizer, and \textbf{AttGF-IQA} is the proposed model trained with GFIs pairs of three corresponding source domains in the GFID training dataset and optimized by meta-learning strategy. Compared with the results of \textbf{w/o meta-learning} and \textbf{AttGF-IQA}, we can observe that the proposed model with meta-learning achieves around 4.0$\%$, 9.8$\%$, 5.4$\%$, 8.0$\%$ improvements on SROCC and 1.2$\%$, 8.8$\%$, 5.6$\%$, 4.0$\%$ improvements on PLCC. Therefore, the proposed model using the meta-learning optimization strategy achieves higher prediction performance in this application.

\begin{table}[t]
\renewcommand{\arraystretch}{1.6}
\tabcolsep0.3cm
  \centering
  \caption{Performance comparisons of the proposed model with different components when training with GFIs pairs of StyleGAN, StarGAN, and PENNet in the GFID training dataset and testing with GFIs of ProGAN in the GFID testing dataset.}
    \begin{tabular}{l|rr}
    \hline
    \hline
        Methods  & \multicolumn{1}{l}{SRCC} & \multicolumn{1}{l}{PLCC} \\
    \hline
    AttGF-IQA w/o CBA & 0.7396&	0.7496 \\
    AttGF-IQA w/o ABA & 0.6923&	0.6922  \\
    AttGF-IQA &  \textbf{0.7728}     &  \textbf{0.7791}\\
    \hline
    \hline
    \end{tabular}%
  \label{tab:ab_components}%
\end{table}%
\subsubsection{Contribution of Different Components}
To evaluate the contributions of different components, we conduct experiments to confirm the benefit from each component of the proposed AttGF-IQA. 
 After removing CBA and ABA components, respectively, the results of those models when training on GFIs pairs of StyleGAN, StarGAN, and PENNet in the GFID training dataset and testing on GFIs of ProGAN in the GFID testing dataset are shown in Table \ref{tab:ab_components}. More specifically, \textbf{AttGF-IQA w/o CBA} means that the AttGF-IQA model removes all CBA modules. \textbf{AttGF-IQA w/o ABA} means that the AttGF-IQA model replaces two AGT of every ABA module with two batch normalization layers. From the table, we can see that the model removing the CBA or ABA module will cause performance degradation.

\section{Conclusion}
\label{sec:conclusion}
In this paper, we focus on studying the GFIs quality assessment from both subjective and objective perspectives. Specifically, we establish the first database GFID for the GFIs quality assessment, which consists of training GFIs pairs with relative quality rankings and testing GFIs with human-annotated scores. Moreover, we design an objective AttGF-IQA model based on the special characteristics of face images and employ the meta-learning optimization strategy to improve generalization ability of the prediction model. Extensive simulation results demonstrate that the proposed AttGF-IQA model achieves higher prediction accuracy and generalization capability on the quality assessment of GFIs than state-of-the-art IQA methods.


%

%


\ifCLASSOPTIONcaptionsoff
  \newpage
\fi



%
%
%
\bibliographystyle{IEEEtran}
\bibliography{strings}

\begin{thebibliography}{10}
\providecommand{\url}[1]{#1}
\csname url@samestyle\endcsname
\providecommand{\newblock}{\relax}
\providecommand{\bibinfo}[2]{#2}
\providecommand{\BIBentrySTDinterwordspacing}{\spaceskip=0pt\relax}
\providecommand{\BIBentryALTinterwordstretchfactor}{4}
\providecommand{\BIBentryALTinterwordspacing}{\spaceskip=\fontdimen2\font plus
\BIBentryALTinterwordstretchfactor\fontdimen3\font minus
  \fontdimen4\font\relax}
\providecommand{\BIBforeignlanguage}[2]{{%
\expandafter\ifx\csname l@#1\endcsname\relax
\typeout{** WARNING: IEEEtran.bst: No hyphenation pattern has been}%
\typeout{** loaded for the language `#1'. Using the pattern for}%
\typeout{** the default language instead.}%
\else
\language=\csname l@#1\endcsname
\fi
#2}}
\providecommand{\BIBdecl}{\relax}
\BIBdecl

\bibitem{GoodfellowG14}
I.~Goodfellow, J.~Pouget-Abadie, M.~Mirza, B.~Xu, D.~Warde-Farley, S.~Ozair,
  A.~Courville, and Y.~Bengio, ``Generative adversarial nets,'' \emph{Adv.
  Neural Inform. Process. Syst.}, pp. 2672--2680, 2014.

\bibitem{WGAN217}
I.~Gulrajani, F.~Ahmed, M.~Arjovsky, V.~Dumoulin, and A.~Courville, ``Improved
  training of wasserstein gans,'' \emph{Adv. Neural Inform. Process. Syst.},
  pp. 5767--5777, 2017.

\bibitem{Progan18}
T.~Karras, T.~Aila, S.~Laine, and J.~Lehtinen, ``Progressive growing of gans
  for improved quality, stability, and variation,'' \emph{Int. Conf. Learn.
  Represent.}, 2018.

\bibitem{StyleGAN19}
T.~Karras, S.~Laine, and T.~Aila, ``A style-based generator architecture for
  generative adversarial networks,'' \emph{IEEE Conf. Comput. Vis. Pattern
  Recog.}, pp. 4401--4410, 2019.

\bibitem{StyleGAN220}
T.~Karras, S.~Laine, M.~Attala, J.~Hellsten, J.~Lehtinen, and T.~Aila,
  ``Analyzing and improving the image quality of stylegan,'' \emph{IEEE Conf.
  Comput. Vis. Pattern Recog.}, pp. 8110--8119, 2020.

\bibitem{StarGAN220}
Y.~Choi, Y.~Uh, J.~Yoo, and J.-W. Ha, ``Starganv2: Diverse image synthesis for
  multiple domains,'' \emph{IEEE Conf. Comput. Vis. Pattern Recog.}, pp.
  8188--8197, 2020.

\bibitem{LahiriP20}
A.~Lahiri, A.~K. Jain, S.~Agrawal, P.~Mitra, and P.~K. Biswas, ``Prior guided
  gan based semantic inpainting,'' \emph{IEEE Conf. Comput. Vis. Pattern
  Recog.}, pp. 13\,696--13\,705, 2020.

\bibitem{ni2020towards}
Z.~Ni, W.~Yang, S.~Wang, L.~Ma, and S.~Kwong, ``Towards unsupervised deep image
  enhancement with generative adversarial network,'' \emph{IEEE Trans. Image
  Process.}, vol.~29, pp. 9140--9151, 2020.

\bibitem{MaskGAN20}
C.-H. Lee, Z.~Liu, L.~Wu, and P.~Luo, ``Maskgan: Towards diverse and
  interactive facial image manipulation,'' \emph{IEEE Conf. Comput. Vis.
  Pattern Recog.}, pp. 5549--5558, 2020.

\bibitem{YangA21}
N.~Yang, Z.~Zheng, M.~Zhou, X.~Guo, L.~Qi, and T.~Wang, ``A domain-guided
  noise-optimization-based inversion method for facial image manipulation,''
  \emph{IEEE Trans. Image Process.}, vol.~30, pp. 6198--6211, 2021.

\bibitem{InterFaceGAN20}
Y.~Shen, C.~Yang, X.~Tang, and B.~Zhou, ``Interfacegan: Interpreting the
  disentangled face representation learned by gans,'' \emph{IEEE Trans. Pattern
  Anal. Mach. Intell.}, 2020.

\bibitem{SSIM04}
Z.~Wang, A.~Bovik, H.~Sheikh, and E.~Simoncelli, ``Image quality assessment:
  From error visibility to structural similarity,'' \emph{IEEE Trans. Image
  Process.}, vol.~13, no.~4, pp. 600--612, 2016.

\bibitem{GMSD14}
W.~Xue, L.~Zhang, X.~Mou, and A.~C. Bovik, ``Gradient magnitude similarity
  deviation: A highly efficient perceptual image quality index,'' \emph{IEEE
  Trans. Image Process.}, vol.~23, no.~2, pp. 684--695, 2014.

\bibitem{ding2020image}
K.~Ding, K.~Ma, S.~Wang, and E.~P. Simoncelli, ``Image quality assessment:
  Unifying structure and texture similarity,'' \emph{IEEE Trans. Pattern Anal.
  Mach. Intell.}, 2020.

\bibitem{RehmanR12}
A.~Rehman and Z.~Wang, ``Reduced-reference image quality assessment by
  structure similarity estimation,'' \emph{IEEE Trans. Image Process.},
  vol.~21, no.~8, pp. 3378--3389, 2012.

\bibitem{GolestanehR16}
S.~Golestaneh and L.~J.Karam, ``Reduced-reference quality assessment based on
  the entropy of dwt coefficients of locally weighted gradient magnitudes,''
  \emph{IEEE Trans. Image Process.}, vol.~25, no.~11, pp. 5293--5303, 2016.

\bibitem{Meta-IQA20}
H.~Zhu, L.~Li, J.~Wu, W.~Dong, and G.~Shi, ``Metaiqa: Deep meta-learning for
  no-reference image quality assessment,'' \emph{IEEE Conf. Comput. Vis.
  Pattern Recog.}, pp. 14\,143--14\,152, 2020.

\bibitem{RankIQA17}
X.~Liu, J.~van~de Weijer, and A.~D. Bagdanov, ``Rankiqa: Learning from rankings
  for no-reference image quality assessment,'' \emph{Int. Conf. Comput. Vis.},
  2017.

\bibitem{MaB21}
J.~Ma, J.~Wu, L.~Li, W.~Dong, X.~Xie, G.~Shi, and W.~Lin, ``Blind image quality
  assessment with active inference,'' \emph{IEEE Trans. Image Process.},
  vol.~30, pp. 3650--3663, 2021.

\bibitem{chen2021no}
B.~Chen, H.~Li, H.~Fan, and S.~Wang, ``No-reference screen content image
  quality assessment with unsupervised domain adaptation,'' \emph{IEEE Trans.
  Image Process.}, 2021.

\bibitem{IS16}
T.~Salimans, I.~Goodfellow, W.~Zaremba, V.~Cheung, A.~Radford, and X.~Chen,
  ``Improved techniques for training gans,'' \emph{Adv. Neural Inform. Process.
  Syst.}, pp. 2234--2242, 2016.

\bibitem{FID17}
M.~Heusel, H.~Ramsauer, T.~Unterthiner, B.~Nessler, and S.~Hochreiter, ``Gans
  trained by a two time-scale update rule converge to a local nash
  equilibrium,'' \emph{Adv. Neural Inform. Process. Syst.}, pp. 6626--6637,
  2017.

\bibitem{Inception16}
C.~Szegedy, V.~Vanhoucke, S.~Ioffe, J.~Shlens, and Z.~Wojna, ``Rethinking the
  inception architecture for computer vision,'' \emph{IEEE Conf. Comput. Vis.
  Pattern Recog.}, pp. 2818--2826, 2016.

\bibitem{Gu20}
S.~Gu, J.~Bao, D.~Chen, and F.~Wen, ``Giqa: Generated image quality
  assessment,'' \emph{Eur. Conf. Comput. Vis.}, pp. 369--385, 2020.

\bibitem{VSI14}
L.~Zhang, Y.~Shen, and H.~Li, ``Vsi: A visual saliency-induced index for
  perceptual image quality assessment,'' \emph{IEEE Trans. Image Process.},
  vol.~23, no.~10, pp. 4270--4281, 2014.

\bibitem{VIF06}
H.~R. Sheikh and A.~C. Bovik, ``Image information and visual quality,''
  \emph{IEEE Trans. Image Process.}, vol.~15, no.~2, pp. 430--444, 2006.

\bibitem{SheikhNo05}
H.~R. Sheikh, A.~C. Bovik, and L.~Cormack, ``No-reference quality assessment
  using natural scene statistics: Jpeg2000,'' \emph{IEEE Trans. Image
  Process.}, vol.~14, no.~11, pp. 1918--1927, 2005.

\bibitem{FerzliA09}
R.~Ferzli and L.~J. Karam, ``A no-reference objective image sharpness metric
  based on the notion of just noticeable blur (jnb),'' \emph{IEEE Trans. Image
  Process.}, vol.~18, no.~4, pp. 717--728, 2009.

\bibitem{MoorthyB11}
A.~K. Moorthy and A.~C. Bovik, ``Blind image quality assessment: From natural
  scene statistics to perceptual quality,'' \emph{IEEE Trans. Image Process.},
  vol.~20, no.~12, pp. 3350--3364, 2011.

\bibitem{BRISQUE12}
A.~Mittal, A.~K. Moorthy, and A.~C. Bovik, ``No-reference image quality
  assessment in the spatial domain,'' \emph{IEEE Trans. Image Process.},
  vol.~21, no.~12, pp. 4695--4708, 2012.

\bibitem{NIQE13}
A.~Mittal, R.~Soundararajan, and A.~C. Bovik, ``Making a ‘completely blind’
  image quality analyzer,'' \emph{IEEE Signal Process. Lett.}, vol.~20, no.~3,
  pp. 209--212, 2013.

\bibitem{MEON18}
K.~Ma, W.~Liu, K.~Zhang, Z.~Duanmu, Z.~Wang, and W.~Zuo, ``End-to-end blind
  image quality assessment using deep neural networks,'' \emph{IEEE Trans.
  Image Process.}, vol.~27, no.~3, pp. 1202--1213, 2018.

\bibitem{DBCNN20}
W.~Zhang, K.~Ma, J.~Yan, D.~Deng, and Z.~Wang, ``Blind image quality assessment
  using a deep bilinear convolutional neural network,'' \emph{IEEE Trans.
  Circuit Syst. Video Technol.}, vol.~30, no.~1, pp. 36--47, 2020.

\bibitem{HyperIQA20}
S.~Su, Q.~Yan, Y.~Zhu, C.~Zhang, X.~Ge, J.~Sun, and Y.~Zhang, ``Blindly assess
  image quality in the wild guided by a self-adaptive hyper network,''
  \emph{IEEE Conf. Comput. Vis. Pattern Recog.}, pp. 3667--3676, 2020.

\bibitem{dipIQ17}
K.~Ma, W.~Liu, T.~Liu, Z.~Wang, and D.~Tao, ``dipiq: Blind image quality
  assessment by learning-to-rank discriminable image pairs,'' \emph{IEEE Trans.
  Image Process.}, vol.~26, no.~8, pp. 3951--3964, 2017.

\bibitem{RankNet05}
C.~Burges, T.~Shaked, E.~Renshaw, A.~Lazier, M.~Deeds, N.~Hamilton, and
  G.~Hullender, ``Learning to rank using gradient descent,'' \emph{Inf. Conf.
  Mach. Learn}, pp. 89--96, 2005.

\bibitem{PENNet19}
Y.~Zeng, J.~Fu, H.~Chao, and B.~Guo, ``Learning pyramid-context encoder network
  for high-quality image inpainting,'' \emph{IEEE Conf. Comput. Vis. Pattern
  Recog.}, pp. 1486--1494, 2019.

\bibitem{LiD18}
H.~Li, S.~J. Pan, S.~Wang, and A.~C. Kot, ``Domain generalization with
  adversarial feature learning,'' \emph{IEEE Conf. Comput. Vis. Pattern
  Recog.}, pp. 5400--5409, 2018.

\bibitem{YangM13}
P.~Yang and W.~Gao, ``Multi-view discriminant transfer learning,''
  \emph{IJCAI}, pp. 1848--1854, 2013.

\bibitem{KhoslaU12}
A.~Khosla, T.~Zhou, T.~Malisiewicz, A.~A. Efros, and A.~Torralba1, ``Undoing
  the damage of dataset bias,'' \emph{Eur. Conf. Comput. Vis.}, pp. 158--171,
  2012.

\bibitem{LiD17}
D.~Li, Y.~Yang, Y.-Z. Song, and T.~M. Hospedales, ``Deeper, broader and artier
  domain generalization,'' \emph{Int. Conf. Comput. Vis.}, pp. 5542--5550,
  2017.

\bibitem{ShankarG18}
S.~Shankar, V.~Piratla, S.~Chakrabarti, S.~Chaudhuri, P.~Jyothi, and
  S.~Sarawagi, ``Generalizing across domains via cross-gradient training,''
  \emph{Int. Conf. Learn. Represent.}, 2018.

\bibitem{ZhouD21}
K.~Zhou, Y.~Yang, Y.~Qiao, and T.~Xiang, ``Domain generalization with
  mixstyle,'' \emph{Int. Conf. Learn. Represent.}, 2021.

\bibitem{LiL18}
D.~Li, Y.~Yang, Y.-Z. Song, and T.~M. Hospedales, ``Learning to generalize:
  Meta-learning for domain generalization,'' \emph{AAAI}, 2018.

\bibitem{SohM20}
J.~W. Soh, S.~Cho, and N.~I. Cho, ``Meta-transfer learning for zero-shot
  super-resolution,'' \emph{IEEE Conf. Comput. Vis. Pattern Recog.}, pp.
  3516--3525, 2020.

\bibitem{IWSSIM11}
Z.~Wang and Q.~Li, ``Information content weighting for perceptual image quality
  assessment,'' \emph{IEEE Trans. Image Process.}, vol.~20, no.~5, pp.
  1185--1198, 2011.

\bibitem{BiSeNet18}
C.~Yu, J.~Wan, C.~Peng, C.~Gao, G.~Yu, and N.~Sang, ``Bisenet: Bilateral
  segmentation network for real-time semantic segmentation,'' \emph{Eur. Conf.
  Comput. Vis.}, pp. 325--341, 2018.

\bibitem{ITU}
``Methodology for the subjective assessment of the quality of television
  pictures,'' \emph{ducument Rec. ITU-R BT.500-13}, 2012.

\bibitem{ni2017esim}
Z.~Ni, L.~Ma, H.~Zeng, J.~Chen, C.~Cai, and K.-K. Ma, ``Esim: Edge similarity
  for screen content image quality assessment,'' \emph{IEEE Trans. Image
  Process.}, vol.~26, no.~10, pp. 4818--4831, 2017.

\bibitem{CBAM18}
S.~Woo, J.~Park, J.-Y. Lee, and I.~S. Kweon, ``Cbam: Convolutional block
  attention module,'' \emph{Eur. Conf. Comput. Vis.}, pp. 3--19, 2018.

\bibitem{U-GAT-IT20}
J.~Kim, M.~Kim, H.~Kang, and K.~Lee, ``U-gat-it: Unsupervised generative
  attentional networks with adaptive layer-instance normalization for
  image-to-image translation,'' \emph{Int. Conf. Learn. Represent.}, 2020.

\bibitem{KumarD11}
N.~Kumar, A.~C. Berg, P.~N. Belhumeur, and S.~K. Nayar, ``Describable visual
  attributes for face verification and image search,'' \emph{IEEE Trans.
  Pattern Anal. Mach. Intell.}, vol.~33, no.~10, pp. 1962--1977, 2011.

\bibitem{SPADE19}
T.~Park, M.-Y. Liu, T.-C. Wang, and J.-Y. Zhu, ``Semantic image synthesis with
  spatially-adaptive normalization,'' \emph{IEEE Conf. Comput. Vis. Pattern
  Recog.}, pp. 2337--2346, 2018.

\bibitem{focal17}
T.-Y. Lin, P.~Goyal, R.~Girshick, K.~He, and P.~Dollar, ``Focal loss for dense
  object detection,'' \emph{Int. Conf. Comput. Vis.}, 2017.

\bibitem{zhu2020deep}
Y.~Zhu, F.~Zhuang, J.~Wang, G.~Ke, J.~Chen, J.~Bian, H.~Xiong, and Q.~He,
  ``Deep subdomain adaptation network for image classification,'' \emph{IEEE
  Trans. Neural Netw. Learn. Syst.}, vol.~32, no.~4, pp. 1713--1722, 2020.

\bibitem{center16}
Y.~Wen, K.~Zhang, Z.~Li, and Y.~Qiao, ``A discriminative feature learning
  approach for deep face recognition,'' \emph{Eur. Conf. Comput. Vis.}, pp.
  499--515, 2016.

\bibitem{FRIQUEE17}
D.~Ghadiyaram and A.~C. Bovik, ``Perceptual quality prediction on authentically
  distorted images using a bag of features approach,'' \emph{J. Vis.}, vol.~17,
  2017.

\bibitem{NIMA18}
H.~Talebi and P.~Milanfar, ``Nima: Neural image assessment,'' \emph{IEEE Trans.
  Image Process.}, vol.~27, no.~8, pp. 3998--4011, 2018.

\bibitem{UNIQUE21}
W.~Zhang, K.~Ma, G.~Zhai, and X.~Yang, ``Uncertainty-aware blind image quality
  assessment in the laboratory and wild,'' \emph{IEEE Trans. Image Process.},
  vol.~30, pp. 3474--3486, 2021.

\bibitem{pytorch}
A.~Paszke, S.~Gross, F.~Massa, A.~Lerer, J.~Bradbury, G.~Chanan, T.~Killeen,
  and Z.~L. et~al{.}, ``Pytorch: An imperative style, high-performance deep
  learning library,'' \emph{Adv. Neural Inform. Process. Syst.}, vol.~32, pp.
  8026--8037, 2019.

\bibitem{Adam}
D.~P. Kingma and J.~Ba, ``Adam: a method for stochastic optimization,''
  \emph{Int. Conf. Learn. Represent.}, 2015.

\bibitem{VQEG}
V.~Q.~E. Group, ``Final report from the video quality experts group on the
  validation of objective models of video quality assessment,'' \emph{VQEG
  meeting}, 2020.

\bibitem{ni2018gabor}
Z.~Ni, H.~Zeng, L.~Ma, J.~Hou, J.~Chen, and K.-K. Ma, ``A gabor feature-based
  quality assessment model for the screen content images,'' \emph{IEEE Trans.
  Image Process.}, vol.~27, no.~9, pp. 4516--4528, 2018.

\end{thebibliography}
%

%
%
%




\end{document}